\documentclass[journal]{IEEEtran}
\usepackage{amsmath,amsfonts}
\usepackage{algorithmic}
\usepackage{algorithm}
\usepackage{array}
\usepackage[caption=false,font=normalsize,labelfont=sf,textfont=sf]{subfig}
\usepackage{textcomp}
\usepackage{stfloats}
\usepackage{url}
\usepackage{verbatim}
\usepackage{graphicx}
\usepackage{cite}

\usepackage{amssymb}
\usepackage{mathtools}
\usepackage{booktabs} 
\usepackage{threeparttable} 
\usepackage{multirow}    
\usepackage{siunitx}        
\usepackage{enumitem}

\begin{document}

\title{NeuroLip: An Event-driven Spatiotemporal Learning Framework for Cross-Scene Lip-Motion-based Visual Speaker Recognition}

\author{Junguang Yao,~\IEEEmembership{Student Member,~IEEE,}
        Wenye Liu,~\IEEEmembership{Member,~IEEE,}\\
        Stjepan Picek,~\IEEEmembership{Senior Member,~IEEE,}
        Yue Zheng,~\IEEEmembership{Member,~IEEE}
\thanks{Manuscript received on March 05, 2026.}
\IEEEcompsocitemizethanks{\IEEEcompsocthanksitem Junguang Yao and  Yue Zheng are with the School of Science and Engineering, The Chinese University of Hong Kong, Shenzhen, Guangdong, 518172, P.R. China (e-mails: junguangyao@link.cuhk.edu.cn; zhengyue@cuhk.edu.cn).

Wenye Liu is an independent researcher (e-mail: wenye.liu@ieee.org).

Stjepan Picek is with the Faculty of Electrical Engineering and Computing University of Zagreb, 10000 Zagreb, Croatia, and Faculty of Science, Radboud University, 6525, XZ Nijmegen, The Netherlands (e-mail: stjepan.picek@ru.nl).

\textit{(Corresponding author: Yue Zheng.)}
\protect\\
} 
\thanks{The authors would like to thank all participants who contributed to the construction of the DVSpeaker dataset. We are grateful to Ying Tao, Xuanyu Chen, Xiaolong Wu, Peichun Hua, Hanxiu Zhang, and Xin Wang for helpful discussions and valuable suggestions on experimental setup and manuscript formatting.}

}

\markboth{Journal of \LaTeX\ Class Files,~Vol.~14, No.~8, August~2021}%
{Shell \MakeLowercase{\textit{et al.}}: A Sample Article Using IEEEtran.cls for IEEE Journals}


\maketitle

\begin{abstract}
Visual speaker recognition based on lip motion offers a silent, hands-free, and behavior-driven biometric solution that remains effective even when acoustic cues are unavailable. Compared to traditional methods that rely heavily on appearance-dependent representations, lip motion encodes subject-specific behavioral dynamics driven by consistent articulation patterns and muscle coordination, offering inherent stability across environmental changes. However, capturing these robust, fine-grained dynamics is challenging for conventional frame-based cameras due to motion blur and low dynamic range. To exploit the intrinsic stability of lip motion and address these sensing limitations, we propose NeuroLip, an event-based framework that captures fine-grained lip dynamics under a strict yet practical cross-scene protocol: training is performed under a single controlled condition, while recognition must generalize to unseen viewing and lighting conditions. NeuroLip features a 1) Temporal-aware Voxel Encoding module with adaptive event weighting, 2) Structure-aware Spatial Enhancer that amplifies discriminative behavioral patterns by suppressing noise while preserving vertically structured motion information, and 3) Polarity Consistency Regularization mechanism to retain motion-direction cues encoded in event polarities. To facilitate systematic evaluation, we introduce \textbf{DVSpeaker}, a comprehensive event-based lip-motion dataset comprising 50 subjects recorded under four distinct viewpoint and illumination scenarios.
Extensive experiments demonstrate that NeuroLip achieves near-perfect matched-scene accuracy and robust cross-scene generalization, attaining over 71\% accuracy on unseen viewpoints and nearly 76\% under low-light conditions, outperforming representative existing methods by at least 8.54\%.
The dataset and code are publicly available at \url{https://github.com/JiuZeongit/NeuroLip}.
\end{abstract}


\section{Introduction}
\label{sec:introduction}

\IEEEPARstart{V}{isual} speaker recognition (VSR) aims to identify individual speakers by analyzing visual cues during speech~\cite{lai2014sparse,he2024lip}. Lip patterns, as inherent biological traits, exhibit stable and discriminative characteristics suitable for personal identification~\cite{tsuchihashi1974studies}. Specifically, speech articulation involves complex lip anatomy and coordinated activation of facial action units~\cite{chowdhury2022lip,newby2025role}, resulting in unique spatiotemporal patterns that encode identity-specific behavior. These patterns are difficult to convincingly imitate or synthesize~\cite{yang2020preventing,zhou2024securing,koch2024one}, offering inherent resistance to impersonation attacks. Consequently, due to its biological nature, non-intrusive acquisition, and inherent insensitivity to environmental acoustic noise, lip-based VSR has emerged as a promising and practical solution for secure access control in applications such as mobile banking recognition, smart surveillance, and biometric forensics~\cite{das2019lip,wright2020understanding}.

Lip patterns are typically categorized into lip texture, lip geometry, and lip motion~\cite{cetingul2006discriminative,wang2012physiological}. Lip texture and geometry are static physiological biometrics, describing lip surface patterns and global lip shape, respectively. Prior studies have mainly focused on such static lip physiological cues and largely relied on a priori knowledge, such as Zernike-based geometric descriptors~\cite{choras2010lip} and sparse coding of local texture patches~\cite{lai2014sparse}, to model appearance or texture-based identity characteristics. In contrast, lip motion is a behavioral biometric characterizing the spatiotemporal dynamics of the lips and perioral muscles during articulation and has been increasingly explored as an identity feature~\cite{luettin1996speaker,cetingul2006discriminative,wang2012physiological,spatiotemporal2011local,wright2020understanding,kuang2023lipauth,koch2024one,chen2025dynamiclip,he2024lip}.
Because motion information is closely coupled with appearance, most existing approaches, especially deep learning-based end-to-end methods~\cite{wright2020understanding,zakeri2024whispernetv2,koch2024one}, fuse static and dynamic information in an entangled manner. As a consequence, the learned identity representations inevitably inherit the sensitivity of appearance cues to imaging conditions~\cite{he2024lip}.

As a result, existing lip-based VSR is sensitive to changes in acquisition conditions~\cite{he2024lip}. In practice, variations in viewing angle and illumination introduce significant distribution shifts: lip geometry may be distorted or partially occluded under profile views, while insufficient lighting blurs fine lip texture. Such mismatches between training and testing data may substantially reduce recognition accuracy. Moreover, in real-world scenarios, users typically provide training samples under controlled conditions with limited training samples, whereas testing often occurs in unconstrained environments. \textit{This motivates a practically important yet underexplored \textbf{cross-scene} problem: learning lip-based identity representations from a single scene during training while maintaining its robustness to unseen variations in viewpoint and illumination at test time}. This setting is particularly relevant when repeatedly collecting data across multiple conditions is costly or impractical, and when the space of possible deployment environments cannot be exhaustively enumerated.

\begin{figure}
\centering
\includegraphics[width=1\linewidth]{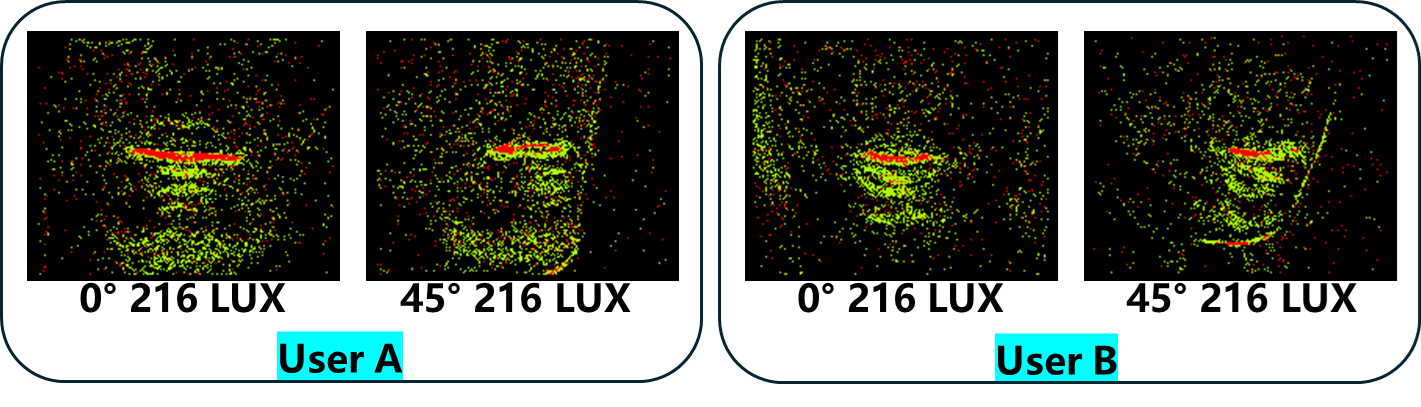} 
\caption{Visualized event frames of two subjects articulating the same digit under different scenarios (viewpoints 0$^\circ$ and 45$^\circ$ under 216 LUX illumination), highlighting subject-specific activation patterns around the lip and chin regions. }
\label{UserAB}
\vspace{-1.4em}
\end{figure}

To our knowledge, systematic investigation of this strict cross-scene setting remains largely absent in the lip-based biometric literature. Prior studies suggest that dynamic behavioral cues are more discriminative and less sensitive to environmental variations than static physiological features~\cite{moreira2022neuromorphic,xiao2014facilitative,wang2012physiological,he2024lip}. Unlike imaging-dependent texture or contour information, temporal motion patterns reflect intrinsic behavioral execution and therefore remain relatively stable across varying viewpoints. However, conventional frame-based imaging sensors often struggle to faithfully capture fine-grained lip motion information during speech; they rely on exposure integration, which leads to inter-frame information loss and susceptibility to illumination variations~\cite{wang2025towards}. 
While alternative sensing modalities such as acoustics~\cite{lu2018lippass} and radar~\cite{yang2025lip} exist, they impose strict environmental constraints and are sensitive to noise. In contrast, event cameras capture intensity changes asynchronously at high temporal resolution, with low quantization error and strong robustness under low illumination~\cite{lenz2022framework,wang2025towards}, making them uniquely suited to capturing fine-grained, rapid lip-motion dynamics during speech. As shown in Fig.~\ref{UserAB}, event-based articulation patterns around the lip and chin regions are subject-specific and consistent across viewpoints, demonstrating the promising viability of using an event camera for robust VSR. 

Despite its practical potential, cross-scene lip-based VSR using event cameras remains highly challenging. Developing deep learning models that generalize from a single acquisition condition to unseen viewpoints and illumination variations presents four key challenges. First, the extracted features must be sufficiently discriminative, exhibiting high inter-subject separability while remaining consistent for the same subject. Second, these features must be robust to scene variations. Viewpoint changes obscure identity-relevant behavioral features, while low illumination degrades signal-to-noise ratio, causing significant feature distribution shifts. Third, it is critical to preserve informative cues from raw event streams. Event data are sparse and asynchronous, and converting them into dense representations for conventional deep learning models often incurs the loss of fine-grained spatial details, temporal continuity, and polarity information that encodes motion directions. Finally, to the best of our knowledge, there are currently no event-based lip motion datasets explicitly designed for cross-scene identity recognition, which further limits systematic investigation of this problem.  

In this paper, we propose NeuroLip, an event-based framework that leverages fine-grained lip motion dynamics for visual speaker recognition. By exploring the asynchronous sensing of event cameras, NeuroLip captures discriminative lip-motion cues from sparse event streams, enabling robust identity recognition across varying environments, even when training is conducted under a single controlled condition.
The main contributions of this work are summarized as follows:
\begin{itemize}
    \item We propose NeuroLip, an event-based framework designed for cross-scene VSR. NeuroLip enables robust generalization to unseen viewpoints and illumination variations when training is performed under a single acquisition condition, which facilitates practical deployment with low-cost user enrollment. 
    \item We introduce DVSpeaker, a new event-based lip-motion dataset explicitly designed to benchmark cross-scene identity recognition. It features 50 subjects recorded under four scenarios with three viewpoints ($0^\circ$, $45^\circ$, and $90^\circ$) and two illumination conditions (12.5 lux and 216 lux) to facilitate systematic evaluation.
    \item We develop a novel representation learning pipeline that bridges raw asynchronous event streams with conventional neural networks (CNNs). By proposing and integrating Temporal-aware Voxel Encoding (TVE), Structure-aware Spatial Enhancer (SSE), and Polarity Consistency Regularization (PCR), our method effectively preserves and exploits fine-grained spatial-temporal behavioral cues.
    \item We conduct extensive experiments to demonstrate the effectiveness and robustness of NeuroLip. It achieves 100\% and 99.97\% accuracy under matched scenes on the self-constructed DVSpeaker and the public DVSLip datasets, respectively, and maintains 71\% accuracy under the strict cross-scene protocol. NeuroLip outperforms 23 representative video-based and event-based methods and exhibits strong data efficiency with as few as 20 training samples per subject.
\end{itemize}

The remainder of this paper is organized as follows.
Section~II reviews related work on lip-based biometrics, event-based classification, and cross-scene identity recognition.
Section~III details the proposed event-based visual speaker recognition framework.
Section~IV describes the design and collection of the event-based lip-motion dataset DVSpeaker.
Section~V presents comprehensive experimental evaluations.
Finally, Section~VI concludes the paper and discusses future research directions.

\section{Related Works}
\label{sec:relatedworks}

\subsection{Lip-based Biometrics}

Early studies established the feasibility of lip biometrics by demonstrating the uniqueness and long-term stability of physiological lip characteristics~\cite{tsuchihashi1974studies}. Subsequent research has primarily focused on static physiological features, with particular emphasis on lip texture and geometric properties. These studies typically relied on strong prior knowledge for feature learning, such as Zernike-based geometric descriptors~\cite{choras2010lip} and sparse coding of local texture patches~\cite{lai2014sparse}. This trend extends to recent deep learning approaches, which often retain explicit priors by integrating handcrafted geometric measurements into probabilistic neural networks~\cite{wrobel2017using} or employing structured texture enhancement guided by Gabor-like filtering in spiking neural networks (SNNs)~\cite{niu2023lip}. 

Lip motion has been explored as an identity cue since the 1990s~\cite{luettin1996speaker}. However, due to the strong coupling between motion and appearance, most approaches jointly model static and dynamic cues. Representative methods include spatiotemporal descriptors based on three orthogonal planes~\cite{spatiotemporal2011local}, video-based authentication pipelines that learn motion and appearance jointly~\cite{wright2020understanding}, and dual-stream architectures that explicitly separate motion and appearance pathways~\cite{zakeri2024whispernetv2}. While studies consistently show that behavioral cues are more discriminative and robust to background interference than physiological features~\cite{moreira2022neuromorphic,xiao2014facilitative,wang2012physiological}, most existing end-to-end models fuse them indiscriminately~\cite{zakeri2024whispernetv2,koch2024one,wright2020understanding}, potentially obscuring the explicit exploitation of identity-relevant motion patterns. Recently, feature disentanglement works~\cite{he2024lip} have advanced this field by explicitly disentangling static identity, dynamic identity, and content features, validating the necessity of focused motion analysis.

While the aforementioned methods rely on conventional frame-based cameras, event cameras use asynchronous sensing with microsecond-level temporal resolution, making them uniquely suited to capturing fine-grained lip motion. Moreira et al.~\cite{moreira2022neuromorphic} presented the first study on event-based VSR, demonstrating that speech-induced facial dynamics captured by event streams can encode discriminative identity information. Their method models short-term facial motion patterns over the full face and validates the effectiveness of dynamic cues for identity recognition. However, the approach is limited to controlled (i.e., fixed viewpoints and illumination) acquisition conditions and relies on full-face information rather than localized lip motion, leading to more data consumption and a higher risk of privacy leakage.

\subsection{Event-based Classification}

Due to the sparse and asynchronous nature of event camera outputs, event-based classification requires dedicated processing pipelines that differ fundamentally from those designed for dense frame-based data. Existing event-based classification methods can be broadly grouped into four representative paradigms. 

One line of work directly processes raw event streams using SNN, which is biologically inspired and naturally suited to sparse temporal signals through spike-based computation. Representative studies include Signed-SGRU for event-based lip reading~\cite{dampfhoffer2024neuromorphic}, which models temporal dynamics in an end-to-end manner, and SpikePoint~\cite{ren2023spikepoint}, which performs efficient action recognition by learning directly from event point clouds without event-to-frame conversion. Despite their conceptual alignment with neuromorphic sensing, SNN-based approaches often suffer from training instability and convergence difficulties due to their discrete computation mechanisms~\cite{tan2022multi}.

Another line of work converts event streams into dense representations (e.g., event frames), enabling the reuse of mature CNNs architectures.
Early work, such as HOTS~\cite{lagorce2016hots}, constructs hierarchical time-surfaces to progressively abstract local spatiotemporal patterns, while TORE volumes~\cite{baldwin2022time} preserve fine-grained temporal ordering by maintaining per-pixel first-in-first-out histories of recent events. Although simple and effective, those representations typically assign equal importance to all events, which can amplify sensor noise, blur informative structures, and weaken the intrinsic sparsity advantage of event data. 

Graph-based methods model events as vertices in spatiotemporal graphs, with edges defined by neighborhood relationships to explicitly encode local dependencies. For example, EGST~\cite{chen2024egst} constructs event graph sequences for efficient gait recognition, while hypergraph-based models~\cite{gao2024hypergraph} further enhance spatiotemporal feature fusion through attention mechanisms. However, graph construction is inherently sensitive to noise, as spurious events may introduce redundant or misleading vertices and edges. 

Finally, point-cloud-based approaches interpret events as 3D points in the spatiotemporal dimension and adopt voxelization or point-based learning strategies from 3D vision. Methods such as~\cite{wang2019space} model event streams as spatiotemporal point clouds and apply PointNet++ for hierarchical feature extraction, while multi-granularity voxel representations are explored in~\cite{tan2022multi} to capture complementary spatial and temporal cues. Nevertheless, voxelization and discretization inevitably introduce quantization errors that may be detrimental to preserving subtle, fine-grained motion patterns.

\subsection{Cross-scene Identity Recognition}
Next, we elaborate on the differences of the following identity recognition protocols: 
\begin{enumerate}[leftmargin=*]
    \item \textbf{Matched-scene}: Training and testing conditions are identical;
    \item \textbf{Cross-scene}:  Training occurs under controlled conditions, while testing takes place in previously \textbf{unseen} scenes.
\end{enumerate}
Comparatively, cross-scene identity recognition is a practical yet challenging scenario that demands robust identity representations invariant to substantial variations in acquisition conditions. Cross-scene identity recognition has been widely studied in face recognition research and is increasingly explored in other biometric research, such as gait-based identification. 

Existing cross-scene face recognition approaches can be broadly categorized into three representative directions: (1) Prior-driven methods: these methods, such as Deep Lambertian Networks~\cite{tang2012deep} and PAM~\cite{tsai2021pam}, rely on explicit physical or geometric priors to mitigate the effects of illumination and viewpoint variations. While effective when their underlying assumptions are satisfied, such works typically struggle to generalize when test conditions deviate from the assumed models. (2) Generative data augmentation-based strategies:  
Methods such as DR-GAN~\cite{tran2017disentangled} and TP-GAN~\cite{huang2017beyond} enhance generalization by synthesizing identity-preserving samples under different poses and lighting conditions. 
However, generating sufficiently realistic and diverse samples remains challenging, and identity fidelity and training stability are not always guaranteed. (3) Geometry-based normalization using 3D morphable models~\cite{blanz2003face,hu2016face}: these works mitigate pose and illumination effects by fitting 2D faces into a unified 3D space. However,  accurate 3D fitting is fragile in practice under large pose variations, occlusion, or low-quality inputs, and the representational capacity of 3D models remains limited. 
Moreover, with the rapid development of deepfake technologies~\cite{he2024lip}, acquiring or forging static facial images has become increasingly inexpensive, raising growing security concerns.

Gait-based identity recognition naturally involves varying viewpoint scenarios, as subjects frequently move through different sensing angles.  However, most gait studies assume that data from diverse conditions are available during the learning phase, which differ fundamentally from cross-scene identity recognition that demands generalization to strictly unseen conditions. To the best of our knowledge, GaitSpike~\cite{tao2024gaitspike} and~\cite{wu2016comprehensive} are among the few works that strictly adopt a cross-scene setting. GaitSpike addresses view variation by leveraging event-based vision and encoding gait dynamics in a floating polar coordinate system centered at the subject’s motion center, which normalizes viewpoint-induced geometric changes and yields view-invariant motion patterns~\cite{tao2024gaitspike}. Wu et al.~\cite{wu2016comprehensive} tackle the same problem from a learning-based perspective by systematically training deep CNNs on multi-view gait data with similarity learning, enabling the model to map gait sequences captured from different camera angles into a shared view-invariant feature space suitable for cross-scene matching. 

In the domain of lip-based identity recognition, the majority of existing works are restricted to matched-scene protocols~\cite{lu2018lippass,yang2025lip,chen2025dynamiclip,kuang2023lipauth} and address scene variations typically by covering all scene conditions during the matched-scene training. For example, DynamicLip~\cite{chen2025dynamiclip} and LipAuth~\cite{kuang2023lipauth} jointly train on diverse illumination, device, and pose conditions, explicitly exposing the model to all expected variations. Alternative sensing modalities, such as acoustic-based Lippass~\cite{lu2018lippass} and radar-based Lip-TWUID~\cite{yang2025lip}, circumvent the visual cross-scene challenge by adopting sensors that are inherently invariant (or less sensitive) to illumination or viewpoints. However, these methods introduce new constraints such as sensitivity to environmental noise or dynamic reflectors. Furthermore, public datasets such as OuluVS2~\cite{anina2015ouluvs2} primarily support multi-view lipreading~\cite{petridis2017end,lee2016multi} rather than cross-scene identification. To date, the stricter cross-scene protocol has not been investigated in lip-based VSR research.  

In this paper, we address this gap by targeting cross-scene recognition, which evaluates robustness to environmental shifts more rigorously than existing matched-scene \cite{lu2018lippass,yang2025lip,chen2025dynamiclip,kuang2023lipauth} protocols. It is particularly important in realistic deployment scenarios where repeatedly collecting data across diverse conditions is costly or impractical, and where future operating environments cannot be exhaustively anticipated. Despite its practical relevance, this challenging configuration remains largely unexplored in current biometric research.

\section{Methodology}

\begin{figure*}[tb]
\centering
\includegraphics[width=0.95\textwidth]{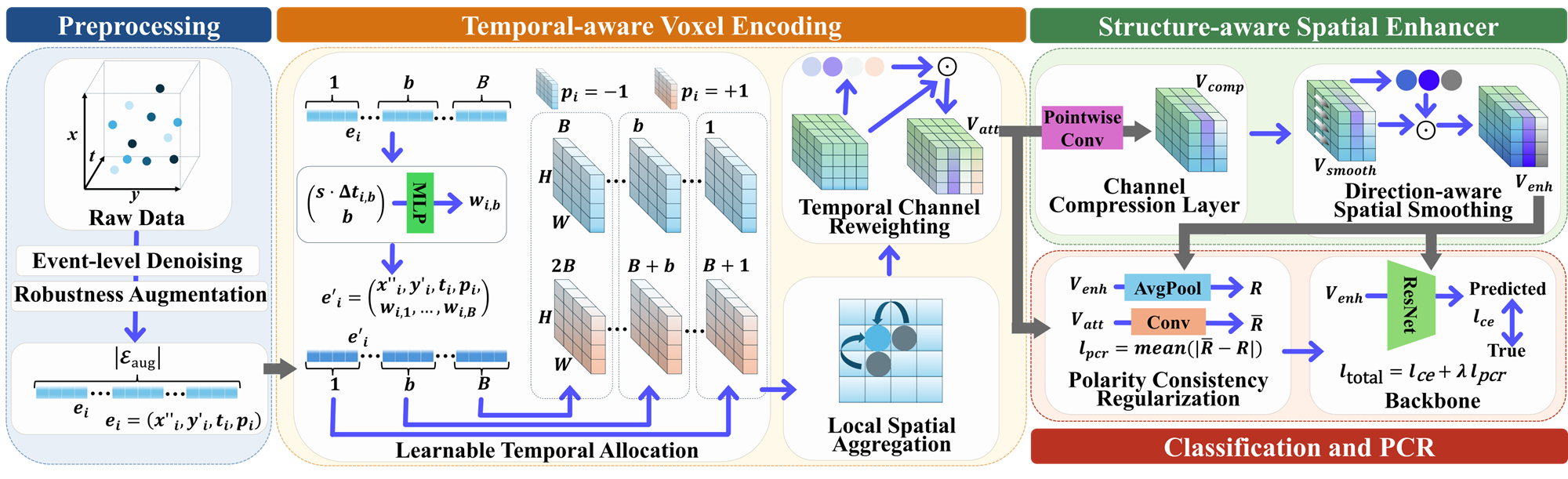}
\caption{An overview of NeuroLip: The pipeline converts raw events into discriminative features via four cascaded stages.}
\label{ProposedMethod}
\end{figure*}

This section details NeuroLip, an end-to-end spatiotemporal learning framework designed for cross-scene lip motion VSR. NeuroLip operates directly on raw event streams and is explicitly designed to address the challenges of event representation mismatch and environmental variance discussed previously. Fig.~\ref{ProposedMethod} depicts the overview of the proposed framework, which is structured into four integrated stages:  
\begin{enumerate}
\item Preprocessing: Performs event-level denoising to suppress spurious events and applies robustness augmentation to mitigate the impact of scene variations and noise. 
\item Temporal-aware Voxel Encoding: Bridges the modality gap by transforming sparse, asynchronous events into dense, information-rich tensors, preserving key spatiotemporal cues via learnable temporal allocation (LTA), local spatial aggregation (LSA), and temporal channel reweighting (TCR).
\item Structure-aware Spatial Enhancer: Refines voxelized features to reduce redundancy and strengthens spatial representations through a channel compression layer (CCL) and direction-aware spatial smoothing (DSS).
\item Classification and PCR: Feeds refined features into a modified ResNet34 backbone and incorporates Polarity Consistency Regularization to explicitly retain motion-direction information carried by event polarities.
\end{enumerate}

The specific design of each component is detailed below.

\subsection{Preprocessing}
Raw event streams inherently contain background noise, especially under low illumination conditions where spurious events surge due to sensor noise. The preprocessing stage aims to improve the signal-to-noise ratio and enhance model generalization through two key operations: event-level denoising and robustness augmentation. Importantly, denoising is applied directly to the raw event stream rather than reconstructed frames to prevent noise amplification during densification. Additionally, augmentation is strictly confined to the training phase to ensure unbiased evaluation.

A raw event stream consisting of $N$ events is defined as
\begin{equation}
\mathcal{E}=\{e_i=(x_i,y_i,t_i,p_i)\mid i=1,2,\ldots,N\},
\end{equation}
where $(x_i,y_i)\in\mathbb{Z}^2$ represent the spatial coordinates of event $e_i$ on the sensor array, $t_i\in\mathbb{R}^+$ denotes the timestamp expressed in microseconds, and 
$p_i\in\{-1,+1\}$ is the polarity indicating a decrease or increase of brightness in the sensing environment.

\subsubsection{Event-level denoising}
Thermal noise or ambient illumination fluctuations~\cite{guo2022low} often trigger isolated events that disrupt the spatiotemporal coherence of motion patterns. Motivated by this observation, we employ a spatiotemporal filter that retains an event only if it possesses at least one distinct neighbor within a local 4-connected spatial neighborhood and a temporal window of $t_f=10\,\text{ms}$. This threshold effectively balances noise suppression with the preservation of valid motion signals.

We define the set of 4-connected spatial neighbors as
\begin{equation}
\mathcal{N}(x_i,y_i)=\{(x_i\pm1,y_i),(x_i,y_i\pm1)\}.
\end{equation}
Accordingly, the denoised stream $\mathcal{E}_{\text{den}}$ is formulated as
\begin{align}
\mathcal{E}_{\text{den}}
=\Big\{e_i\in\mathcal{E}\ \Big|\ 
&\exists\, e_j\in\mathcal{E},\ j\neq i,\ (x_j,y_j)\in\mathcal{N}(x_i,y_i),\notag\\
&|t_j-t_i|<t_f\Big\}.
\end{align}
Notably, we do not impose a polarity constraint on $e_j$: neighbors of either polarity serve as valid evidence of local spatiotemporal activity. 

\subsubsection{Robustness augmentation}
To mitigate overfitting and improve robustness against viewpoint and illumination variations, we apply three augmentation strategies exclusively during the training phase.
Unless otherwise specified, these operations are applied sequentially: spatial translation, horizontal mirroring, and event sparsification.

\paragraph{Spatial translation}
We introduce random spatial translation to simulate a minor viewpoint shift. Spatial offsets are sampled from a uniform distribution $\mathcal{U}(-20,20)$ over the interval $[-20,20]$, i.e.,  $\Delta x,\Delta y\sim\mathcal{U}(-20,20)$. Accordingly, the spatially shifted event stream $\mathcal{E}_{\text{shift}}$ retains only those events that fall within the valid sensor resolution (W, H): 
\begin{align}
x_i'&=x_i+\Delta x,\quad y_i'=y_i+\Delta y,\\
\mathcal{E}_{\text{shift}}
&=\{(x_i',y_i',t_i,p_i)\mid 0\le x_i'<W,\ 0\le y_i'<H\}.
\end{align}

\paragraph{Horizontal mirroring}
We apply horizontal mirroring with a probability of $0.5$ to simulate symmetric lip motions. The coordinates are transformed as:
\begin{equation}
x_i''=
\begin{cases}
W-1-x_i', & \text{if }\mathrm{u}<0.5,\\
x_i', & \text{otherwise},
\end{cases}
\end{equation}
where $u\sim \mathcal{U}(0,1)$ is a random variable sampled from a uniform distribution. This yields the mirrored set $\mathcal{E}_{\text{flip}}=\{(x_i'',y_i',t_i,p_i)\}$. Any events falling outside the sensor boundaries post-transformation are discarded.

\paragraph{Event sparsification}
To simulate the sparsity characteristic of low-light conditions, we randomly discard a portion of events from the processed stream $\mathcal{E}_{\text{flip}}$. We sampled a drop ratio $r\sim\mathcal{U}(0.4,0.7)$ and determine the target count $ T=\lfloor N'(1-r)\rfloor $, where $N'=|\mathcal{E}_{\text{flip}}|$. The final augmented stream $\mathcal{E}_{\text{aug}}$ is obtained by uniformly sampling $T$ events from $\mathcal{E}_{\text{flip}}$ without replacement: 
\begin{equation}
\mathcal{E}_{\text{aug}}\subset \mathcal{E}_{\text{flip}}, \quad |\mathcal{E}_{\text{aug}}|=T.
\end{equation}
To prevent excessive information loss that might erase critical motion cues, this sparsification step is applied to only 30\% of the training samples. We tune the sparsification probability on the validation split and find that applying it to 30\% of training samples yields the best trade-off: higher probabilities over-sparsify inputs and degrade performance, while lower probabilities provide insufficient robustness gain.

\subsection{Temporal-aware Voxel Encoding}
This section details the proposed TVE, which transforms sparse, asynchronous events into a dense voxel tensor. This encoding is explicitly designed to preserve fine-grained temporal dynamics, polarity information, and local spatial continuity, thereby maximizing information extraction from the raw event data.

\subsubsection{Time normalization}
To eliminate dependency on the variable recording duration, we first normalize the event timestamps to the range $[0,1]$:
\begin{equation}
t_i^{\text{norm}}=\frac{t_i-t_1}{t_T-t_1}.
\end{equation}
where $t_1$ and $t_T$ denote the first ($i=1$) and last ($i=T$) timestamps in the sequence, respectively.

\subsubsection{Learnable temporal allocation}
Voxelization is a widely used event representation technique that maps continuous asynchronous events into discrete grid-based structures. Standard voxelization typically discretizes the time domain into fixed temporal bins, accumulating event polarity counts within each bin to form a dense representation. However, most existing approaches discard fine-grained temporal information by hard-assigning each event to a single specific temporal bin, thereby quantizing the continuous timestamps. To address this quantization loss, we adopt learnable temporal allocation. Instead of a hard assignment, LTA allows each event to contribute to multiple temporal bins via learnable weights, thereby preserving fine-grained temporal dynamics.
Let $B$ denote the number of temporal bins. We first compute the temporal offset of each event $e_i$ with respect to the starting point of each bin:  
\begin{equation}
\Delta t_{i,b}=t_i^{\text{norm}}-\frac{b-1}{B}, \quad b \in\{1,\ldots,B\}.
\end{equation}
Next, we determine the allocation weights $w_{i,b}$, representing the contribution of event $e_i$ to bin $b$, using a lightweight two-layer multilayer perception (MLP):
\begin{equation}
w_{i,b}=MLP\!\left(s\cdot\Delta t_{i,b},\, b\right),
\end{equation} where $s\in\mathbb{R}$ is a learnable scaling parameter that modulates the sensitivity to temporal offsets.
Specifically, a larger $s$ amplifies small offsets, resulting in sharper temporal responses, where a smaller $s$ encourages smoother weight distribution across temporal bins. The allocation weights are produced by a lightweight two-layer MLP that takes $(s\Delta t_{i,b},\, b)$ as input, uses a 32-unit hidden layer with ReLU, and outputs a scalar weight. This pointwise regressor is chosen to keep the encoder parameter- and compute-efficient while remaining flexible enough to approximate common temporal interpolation kernels. Note that $\text{MLP}$ can also be implemented using other pointwise mapping methods such as CNN-based encoders. The explicit inclusion of the bin index $b$ enables bin-dependent temporal allocation,
allowing different temporal bins to emphasize distinct phases of motion dynamics without introducing any spatial coupling.

Based on the computed allocation weights $w_{i,b}$ for all events, we perform LSA to construct the dense voxel tensor. We spatially aggregate the weighted contribution of events into $2B$ feature maps, explicitly separating positive and negative polarities. Specifically, for each polarity $p\in\{+1, -1\}$ and temporal bin $b$, the pixel value at spatial coordinate (x, y) is obtained by summing the allocation weights of all valid events located at that position. This process yields a final dual-polarity voxel tensor $V\in\mathbb{R}^{2B\times H\times W}$. By using learned weights rather than binary counts, this aggregation strategy embeds precise sub-bin temporal dynamics directly into the feature magnitude, preventing the loss of fine-grained motion cues typical in standard voxelization.

\subsubsection{Local spatial aggregation}
Event streams exhibit strong spatial correlations across adjacent pixels, where meaningful structures such as lip contours arise from local continuity rather than isolated pixels.
To exploit such dependencies, we incorporate an LSA during voxelization. Specifically, we update each voxel by accumulating values from its immediate 2-connected neighborhood, thereby reinforcing structural connectivity:
\begin{equation}
V_{c, y, x} =V_{c, y, x}+ V_{c, y+1, x}+ V_{c, y, x+1},
\end{equation}
where $c\in\{1, \cdots, 2B\}$ denotes the channel index and out-of-bound indices are discarded.

By aggregating only a targeted subset of neighbors (asymmetric offsets), this design effectively captures local spatial continuity while avoiding the excessive blurring and redundant accumulation associated with a larger, symmetric smoothing kernel.

\subsubsection{Temporal channel reweighting}
Different temporal bins may contribute unequally to identity discrimination. To address this, we introduce TCR to adaptively emphasize informative temporal bins while suppressing redundant or noisy ones.

First, we perform Global Average Pooling (GAP) over each $W\times H$ feature map, compressing the spatial dimensions of the voxel tensor $\boldsymbol{V} \in \mathbb{R}^{2B \times H \times W}$ into a channel descriptor $z\in \mathbb{R}^{2B}$ by aggregating global spatial context. Since the channel dimension corresponds to temporal bins and polarities, this descriptor $z$ effectively represents an ordered temporal sequence of the lip motion intensity. Next, we apply a 1D convolution $\text{Conv1d}(\cdot)$ (kernel size of 1 in our experiments) to capture local interactions between adjacent temporal bins, followed by a Sigmoid activation to generate attention calibration weights: 
\begin{equation}
\boldsymbol{a}_{\text{ch}} =
\sigma\!\left(
\text{Conv1d}\!\left(
\text{GAP}(\boldsymbol{V})\right)
\right) \in \mathbb{R}^{2B},
\end{equation}
where $\sigma$ denotes the Sigmoid function. Finally, the original voxel features are recalibrated via channel-wise multiplication $\odot$: 
\begin{equation}
\boldsymbol{V}_{\text{att}} = \boldsymbol{V} \odot \boldsymbol{a}_{\text{ch}}.
\end{equation}
By learning global correlations among polarity-aware temporal bins, TCR effectively highlights critical motion phases without altering spatial resolution or introducing heavy temporal convolutions.

\subsection{Structure-aware Spatial Enhancer}
Although $\boldsymbol{V}_{\text{att}}$ is dense, it may still retain channel redundancy and localized spatial anomalies caused by residual noise.
The SSE addresses these issues by refining the spatial representations through two sequential steps: channel compression and direction-aware spatial smoothing.

\subsubsection{Channel compression layer}
The voxel tensor $\boldsymbol{V}_{\text{att}}$ produced by TVE contains $2B$ channels, explicitly separates temporal bins and polarities, which may introduce inter-channel redundancy. To distill a compact yet expressive representation for spatial enhancement, we first employ a CCL.

Specifically, we apply a pointwise $1\times1$ convolution $\text{Conv}_{1\times1}(\cdot)$ to  linearly project the high-dimensional channel features into a lower-dimensional space:
\begin{equation}
\boldsymbol{V}_{\text{comp}} = \text{Conv}_{1\times1}(\boldsymbol{V}_{\text{att}})\in\mathbb{R}^{C\times H\times W}.
\end{equation}
This operation reduces the channel dimensionality from $2B$ to $C$, facilitating the fusion of information across different temporal bins and polarities. Empirically, this reduction achieves an optimal trade-off between preserving discriminative lip-motion cues and suppressing task-irrelevant noise. Distinct from the attention-based TCR module, this layer focuses on per-pixel channel mixing and dimensionality reduction, thereby improving computational efficiency for the subsequent spatial smoothing stage.

\subsubsection{Direction-aware spatial smoothing}
In lip motion analysis, vertically structured cues such as lip opening and closing boundaries are typically more informative than purely horizontal variations. To preserve these critical vertical structures while suppressing high-frequency horizontal noise, we introduce DSS. 

Specifically, we apply a depthwise 1D convolution along the horizontal axis of the feature map. By operating independently on each channel, this step smooths horizontal fluctuations without blurring the informative vertical structural continuity. The smoothed features are then normalized and activated to stabilize training and enhance non-linearity: 
\begin{equation}
\boldsymbol{V}_{\text{smooth}}
=\text{ReLU}\!\left(\text{BatchNorm}\!\left(\text{DWConv1d}(\boldsymbol{V}_{\text{comp}})\right)\right),
\end{equation}
where $\text{DWConv1d}(\cdot)$ denotes a depthwise 1D convolution with a kernel size of $3$ and zero-padding of size $1$, applied exclusively along the horizontal dimension. We set the kernel size to $3$ with padding $1$ to perform the smallest effective local smoothing while preserving the feature map resolution. This helps suppress high-frequency horizontal noise without excessively blurring the vertically structured lip boundaries, and keeps the operation lightweight.

Following spatial smoothing, we employ a channel-wise attention mechanism to adaptively recalibrate the feature channels based on their enhanced structural saliency. Similar to the TCR module, we use $\text{GAP}$ followed by a 1D convolution to generate channel-wise attention weights:
\begin{align}
\boldsymbol{a}_{\text{sp}}
&=\sigma\!\left(\text{Conv1d}\!\left(\text{GAP}(\boldsymbol{V}_{\text{smooth}})\right)\right),\\
\boldsymbol{V}_{\text{enh}}
&=\boldsymbol{V}_{\text{smooth}}\odot \boldsymbol{a}_{\text{sp}}.
\end{align}
Here, $\boldsymbol{a}_{\text{sp}}\in \mathbb{R}^{C}$ represents the learned importance of each smoothed channel. This results in the final refined tensor $\boldsymbol{V}_{\text{enh}}\in\mathbb{R}^{C\times H\times W}$, which effectively encodes spatiotemporal features with enhanced structural saliency.

\subsection{Classification and PCR}
This stage performs identity classification using a modified ResNet34 backbone, augmented with PCR. While deep networks excel at extracting semantic features, repeated spatial aggregation and channel mixing in SSE risk dilute the low-level polarity cues that encode precise motion direction. PCR is designed to counter this by enforcing enhanced feature representations that retain polarity distributions consistent with the original SSE input.

\subsubsection{Polarity consistency regularization}
To estimate the preserved polarity information, we employ a lightweight convolutional reconstruction head that maps the enhanced features $V_\emph{enh}$ back to a dual-channel polarity map $\boldsymbol{R}\in\mathbb{R}^{2\times H\times W}$, corresponding to negative and positive polarities. To ensure sufficient representational capacity for this mapping, we utilize a two-layer CNN structure: features are first projected to a higher-dimensional space to capture richer interactions, and then mapped to the target 2-channel polarity output:
\begin{align}
\boldsymbol{R} &= \text{Conv2d}(\text{ReLU}\!\left(\text{Conv2d}(\boldsymbol{V}_{\text{enh}})\right)),
\end{align}
where $\text{Conv2d}$ denotes a standard 2D convolution. 
We adopt a two-layer CNN as a minimal reconstruction head: a single convolution is often insufficient to recover polarity patterns after channel mixing and smoothing, while deeper heads add unnecessary parameters and may overfit. The expansion-compression design provides enough capacity to model local interactions and then project to the required 2-channel polarity map. This expansion-compression design enhances the head’s ability to recover fine-grained polarity details from the compressed feature space. 

We derive the reference tensor $\bar{\boldsymbol{R}}\in\mathbb{R}^{2\times H\times W}$ by averaging the polarity-specific temporal channels from the initial voxel encoding $\boldsymbol{V}_{\text{att}}$. Let $\alpha \in \{0,1\}$ index the negative and positive polarity partitions of $\boldsymbol{V}_{\text{att}}$, respectively. The reference map is computed as: 
\begin{equation}
\bar{\boldsymbol{R}}[\alpha,y,x]
=
\frac{1}{B}
\sum_{i=1}^{B}
\boldsymbol{V}_{\text{att}}[\alpha B+i, y, x].
\end{equation}

Since absolute event magnitude can vary significantly with illumination, we focus on preserving the spatial distribution of polarity rather than absolute intensity. We apply spatial normalization to both the reconstructed and reference polarity maps: 
\begin{align}
\boldsymbol{R}_{\text{norm}}[\alpha,y,x]
&=\frac{\boldsymbol{R}[\alpha,y,x]}{\sum_{y,x}\boldsymbol{R}[\alpha,y,x]},\\
\bar{\boldsymbol{R}}_{\text{norm}}[\alpha,y,x]
&=\frac{\bar{\boldsymbol{R}}[\alpha,y,x]}{\sum_{y,x}\bar{\boldsymbol{R}}[\alpha,y,x]}.
\end{align}
The PCR loss is then defined as the $\ell_1$ distance between these normalized distributions:
\begin{equation}
l_{\text{pcr}}
=\frac{1}{2HW}\sum_{\alpha,y,x}
\left|
\boldsymbol{R}_{\text{norm}}[\alpha,y,x]
-
\bar{\boldsymbol{R}}_{\text{norm}}[\alpha,y,x]
\right|.
\end{equation}
By minimizing $l_{\text{pcr}}$, the network is encouraged to preserve polarity-aware motion information in the learned features.

\subsubsection{ResNet34 backbone and total loss}
ResNet34 was originally designed for 3-channel RGB images and employs aggressive early downsampling, which risks discarding the fine-grained motion cues critical for event-based analysis. To adapt the backbone for our enhanced event representations, we introduce two key modifications: (1) We replace the initial $7\times7$ stride-2 convolution with a $3\times3$ stride-1 convolution, configured to accept the $C$ channel input from the SSE module; (2) We replace the stride-2 max pooling with an identity mapping. These changes preserve the spatial resolution of sparse but discriminative motion patterns in the early stage of the network.

The framework is trained end-to-end by jointly optimizing the classification and polarity preservation. The total objective function is defined as:
\begin{equation}
\label{TotalLoss}
l_{\text{total}}=l_{\text{ce}}+\lambda\,l_{\text{pcr}},
\end{equation}
where $l_{\text{ce}}$ is the cross-entropy classification loss and $\lambda$ is a balancing hyperparameter. We empirically set $\lambda=0.05$ to ensure that the PCR regularization guides feature learning without dominating the primary classification task.

\section{The DVSpeaker Dataset}
\label{sec:dvspeaker}

To address the lack of suitable resources for cross-scene analysis, we introduce DVSpeaker, a new and comprehensive event-based lip-motion dataset. DVSpeaker is explicitly designed to support identity recognition under challenging real-world scenarios, incorporating systematic variations in viewpoints and illumination conditions. The design rationale, acquisition setup, processing pipeline, and statistical characteristics of DVSpeaker are detailed below. 

\subsection{Design Rationale}
Existing lip-related datasets generally follow three paradigms: digit-based, word-based, or sentence-based. The digit paradigm, widely used in identity recognition tasks, requires subjects to articulate digits with consistent and simple phonetic structures, and has been adopted by datasets such as XM2VTS~\cite{messer1999xm2vtsdb}, Tulips~\cite{movellan1994visual}, qFace, and FAVLIP~\cite{wright2020understanding}. The word paradigm serves both identity recognition and lip-reading tasks, including MVGL-AVD~\cite{erzin2005multimodal}, LRW~\cite{chung2016lip}, and DVSLip~\cite{tan2022multi}. Sentence-based datasets, such as CREMA-D~\cite{cao2014crema}, TIMIT~\cite{zue1990speech}, GRID~\cite{cooke2006audio}, and GRID-CCP~\cite{koch2024one}, are primarily employed for emotion analysis and lip-reading. 

For DVSpeaker, we adopt a digit-based paradigm (0--9) rather than words or sentences. This design choice is driven by three key reasons aimed at minimizing linguistic confounding factors, controlling phonetic complexity, and expanding real-world application scenarios. First, sentence-level and long-word utterances introduce coarticulation effects~\cite{Benedikt2010}, which may blur subject-specific lip-motion patterns, whereas digits exhibit relatively independent and well-defined articulation boundaries. Second, due to the interpretability of deep learning models, complex linguistic content may introduce confounding factors unrelated to identity, such as word-specific lip shapes or temporal intervals~\cite{he2024lip}. Using digits with uniform phonetic complexity encourages deep learning models to focus on identity-related motion cues. Last but not least, digit-based protocols also align with practical security scenarios, such as ID cards, phone numbers, and doorplates~\cite{koch2024one,kuang2023lipauth,lu2018lippass,yang2020preventing,zhou2024securing}. 

\subsection{Acquisition Setup}

\begin{figure}[tb]
\centering
\includegraphics[width=1\linewidth]{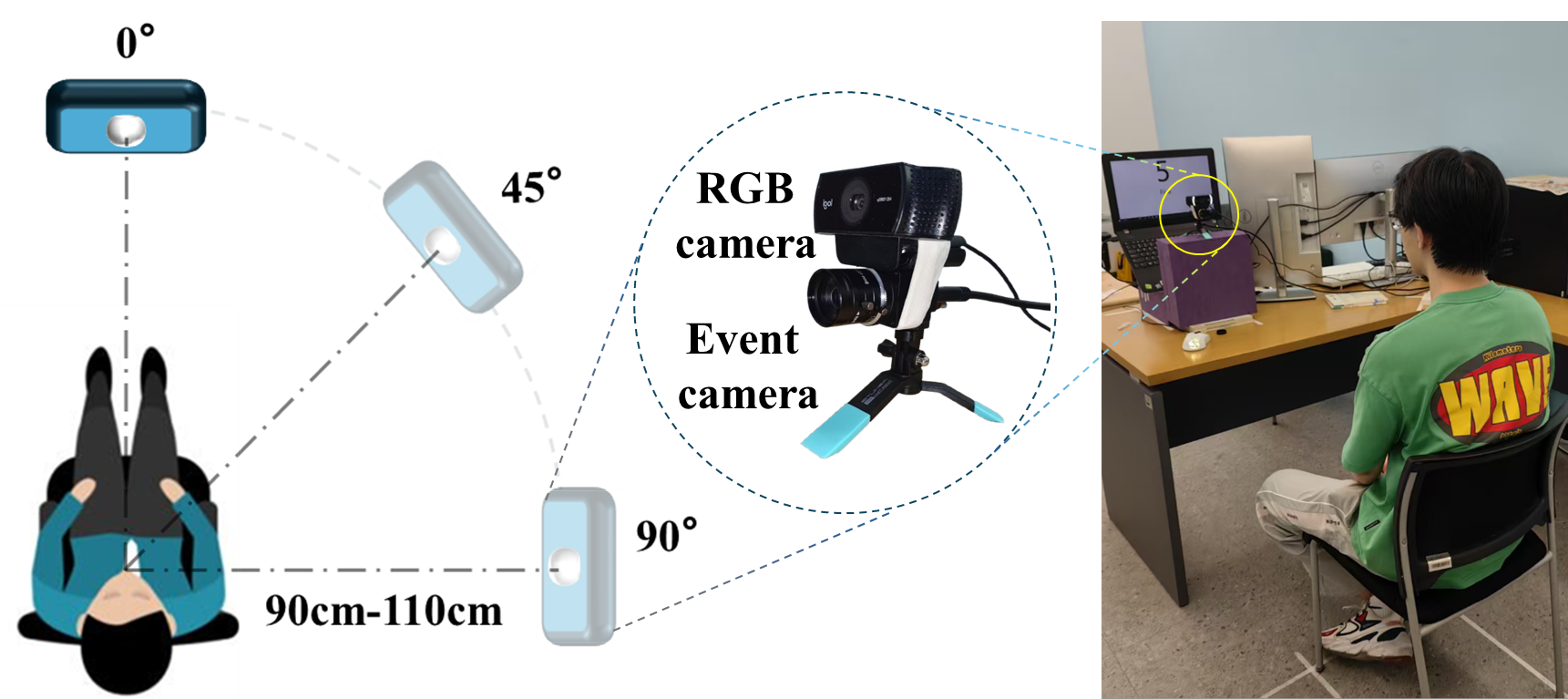} 
\caption{Acquisition setup using a synchronized dual-camera (event and RGB) system. Viewpoints ($0^\circ$, $45^\circ$, $90^\circ$) are adjusted by repositioning the cameras along a horizontal arc while maintaining a constant distance from the stationary subject. Illumination intensity is controlled via indoor lighting. }
\label{ExperimentProcedure}
\vspace{-1.4em}
\end{figure}

To construct DVSpeaker, we recruited 50 healthy participants (25 males and 25 females) with a mean age of $24.0 \pm 2.9$ years. All participants had normal or corrected-to-normal vision and reported no history of speech or lip-motor impairments. The study was approved by the Ethics Review Board of The Chinese University of Hong Kong, Shenzhen (Ethics ID: CUHKSZ-D-20250005), and written informed consent was obtained from all participants prior to data collection. Participants received appropriate compensation upon completion of the experimental procedure.

A dual-camera system is utilized to capture synchronized visual data. We employ a Prophesee EVK4 event camera (1280$\times$720 resolution) to record high-temporal-resolution lip dynamics, alongside a Logitech C922 color camera (1920$\times$1080 resolution, 60 FPS) to provide RGB reference frames. The two image sensors are rigidly mounted together to ensure synchronized data capture of the same scene. As illustrated in Fig.~\ref{ExperimentProcedure}, participants were seated at a distance of 90--110 cm from the cameras and instructed to maintain a stable posture. Visual stimuli were displayed on a monitor positioned behind the cameras, presenting randomly ordered digits (0--9) at 3-second intervals. Subjects were asked to articulate each digit naturally and clearly. To mitigate fatigue and maintain articulatory consistency, a mandatory rest period of at least one minute was enforced after every 50 trials.

Data collection was conducted under four distinct acquisition conditions to introduce systematic variations in viewpoint and illumination. Ambient light intensity was quantified using a TASI TA630A light meter. Let SI denote the ``Sufficient Illumination" ($\approx$216~lux) and II denote ``Insufficient illumination" ($\approx$12.5~lux). Our dataset consider four scenario: SI-$0^\circ$, SI-$45^\circ$, SI-$90^\circ$ and II-$0^\circ$, which represent frontal view ($0^\circ$), 45-degree view ($45^\circ$) and 90-degree view ($90^\circ$) under sufficient illuminations ($\approx$216~lux), and low-light frontal view ($0^\circ$) under insufficient illumination ($\approx$12.5~lux). For each condition, we collected 100 valid samples per participant, ensuring a comprehensive and balanced dataset for cross-scene evaluation. 

\subsection{Event Data Curation and Annotation}
 To facilitate manual annotation and quality verification, the raw events captured by the EVK4 sensor were first converted into visualization frames using the Metavision SDK (version 5.0). Using these frames, the lip regions of interest containing the lips were manually annotated, and each sample was spatially cropped to a resolution of $200 \times 160$ pixels, retaining only the events falling within this window. For accurate temporal segmentation, we also annotated the onset timestamp of each articulation. 
Based on these timestamps and the annotations, we extracted 3-second event segments corresponding to individual digit articulations from the continuous raw event streams. Finally, each segment was assigned its specific digit label (0--9), yielding a temporally aligned and well-structured dataset suitable for supervised learning.

\subsection{Dataset Statistics}

\begin{figure}[tb]
\centering
\includegraphics[width=1\linewidth]{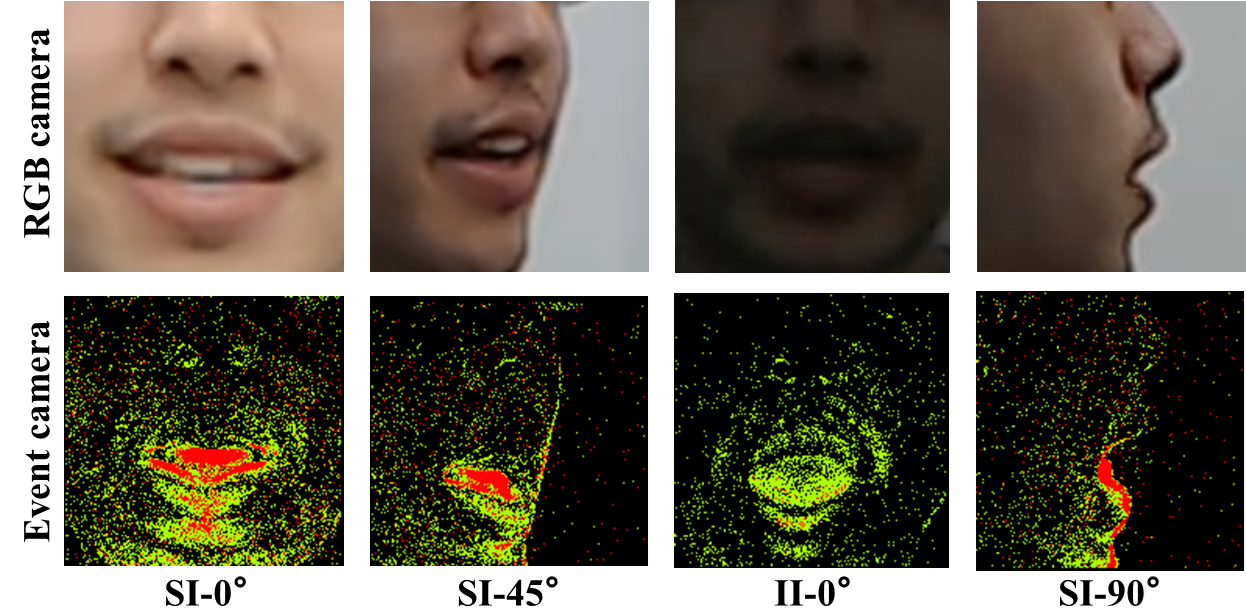} 
\caption{Representative samples from DVSpeaker across four experimental conditions with varying viewpoints and illumination. Top: RGB reference frames; Bottom: Corresponding event streams (visualized as accumulated frames).}
\label{SomeSamples}
\vspace{-1.4em}
\end{figure}

Fig.~\ref{SomeSamples} presents a few visual samples of DVSpeaker captured under different viewpoints and illumination conditions. These examples illustrate the diversity of lip motion dynamics and demonstrate the effectiveness of our curation pipeline.
The finalized dataset comprises 20,000 valid samples, with each of the 50 participants contributing 400 samples. Specifically, each participant provides 40 samples per digit (0--9), which are evenly distributed across the four experimental conditions. All samples are standardized to a spatial resolution of $200 \times 160$ pixels and a temporal duration of 3 seconds, ensuring consistency across model training and evaluation.

\section{Experimental Analysis}

In this section, we conduct comprehensive experiments to evaluate NeuroLip. First, we report the matched-scene and cross-scene performance, followed by a comparative analysis against existing approaches. We then investigate the model’s adaptability by extending our cross-scene protocol to a few-shot domain adaptation setting. Finally, we present detailed ablation studies to analyze the impact of essential components, hyperparameters (PCR strength $\lambda$, compressed channel dimension $C^\prime$), spoken contents, and training sample size.

\subsection{Datasets and Implementation Details}
The proposed system is evaluated using two datasets: our self-constructed cross-scene dataset DVSpeaker and the public matched-scene dataset DVSLip~\cite{tan2022multi}. As introduced in Sec.~\ref{sec:dvspeaker}, DVSpeaker comprises 20,000 3-second event stream samples, evenly contributed by 50 speakers articulating digits 0--9. Samples are captured under four experimental conditions: SI-$0^\circ$, SI-$45^\circ$, SI-$90^\circ$, and II-$0^\circ$, and each has been standardized to a spatial resolution of 200$\times$160 pixels. As for DVSLip, it is a public word-based lip dataset captured under a single controlled environment, containing recordings of 40 speakers pronouncing 100 English words (five repetitions each), with durations of 0.2-1.4 seconds and a resolution of $128\times128$. Notably, while both datasets supported matched-scene evaluation, DVSpeaker is currently the only benchmark suitable for cross-scene lip motion VSR. 

All experiments are implemented in PyTorch on a NVIDIA GeForce RTX 4090 GPU (CUDA 12.8). The network is trained for 30 epochs\footnote{30 epochs serve as a sufficient upper bound for convergence under our setting.} using the Adam optimizer with a learning rate of $10^{-4}$ and a batch size of 8. An exponential learning-rate scheduler with a decay factor of 0.5 is applied every 10 epochs. To ensure rigorous evaluation, the checkpoint achieving the lowest validation loss is selected for evaluation. 
Our evaluation protocols are defined as follows. For matched-scene evaluation, all samples in the dataset are partitioned into training, validation, and testing sets with a 6:2:2 ratio. For cross-scene evaluations (DVSpeaker only), testing samples are strictly unseen during model development. Specifically, all samples from the target test scenario are reserved exclusively for evaluation, while data from the designated source condition is partitioned into training and validation sets at an 8:2 ratio. This protocol rigorously simulates a real-world deployment where the target environment is unseen during model development.

\subsection{Matched-scene and Cross-scene Performance}

We first evaluate NeuroLip under standard matched-scene protocols, where training and testing data are drawn from the same scenario (distribution). Table~\ref{tab:our_method_datasets} reports the matched-scene recognition accuracy on both DVSpeaker and DVSLip. On DVSpeaker, NeuroLip achieves 100\% accuracy across all four acquisition conditions. Similarly, on DVSLip, our method attains a near-perfect accuracy of 99.97\%. These results demonstrate that NeuroLip effectively learns highly discriminative identity features from lip motion, exhibiting robust performance across diverse dataset configurations and speaking contents. 

\begin{table}[htbp]
\centering
\footnotesize
\setlength{\tabcolsep}{4pt}
\caption{Matched-scene Performance on DVSpeaker and DVSLip}
\label{tab:our_method_datasets}
\begin{threeparttable}
\begin{tabular}{l*{5}{S[table-format=3.2]}}
\toprule
\multicolumn{1}{c}{\textbf{Dataset}} & \multicolumn{1}{c}{\textbf{DVSLip~\cite{tan2022multi}}} & \multicolumn{4}{c}{\textbf{DVSpeaker}} \\
\cmidrule(lr){3-6}
& & \textbf{SI-$0^\circ$} & \textbf{SI-$45^\circ$} & \textbf{SI-$90^\circ$} & \textbf{II-$0^\circ$} \\
\midrule
Accuracy (\%) & 99.97 & 100.00 & 100.00 & 100.00 & 100.00 \\
\bottomrule
\end{tabular}
\end{threeparttable}
\end{table}

To further assess NeuroLip’s robustness under domain shifts, we conduct a cross-scene evaluation on DVSpeaker. Specifically, the model is trained exclusively on the SI-$0^\circ$ condition and evaluated directly on the unseen SI-$45^\circ$, SI-$90^\circ$, and II-$0^\circ$ conditions. We select the SI-$0^\circ$ condition for training because frontal views under sufficient illumination provide the most complete and stable lip-motion observations, mimicking standard controlled enrollment scenarios. As shown in Table~\ref{tab:baseline_train10}, NeuroLip achieves an accuracy of 71.68\% on SI-$45^\circ$ and 75.94\% on II-$0^\circ$ under cross-scene settings. While a performance drop is observed compared to the matched-scene baseline (100\% in Table~\ref{tab:our_method_datasets}), the model still retains significant discriminative capability in these challenging settings. Given the cross-scene protocol, these results far exceed random guessing (2\%), suggesting that the learned representations are not solely reliant on low-level geometry- or illumination-dependent appearance cues, but successfully encode stable, identity-specific behavioral characteristics. We note that performance on the extreme SI-90$^\circ$ condition is limited (3.64\%), indicating that orthogonal viewpoint changes remain a formidable challenge for models due to severe self-occlusion. 
However, NeuroLip still represents a substantial leap forward, significantly outperforming existing approaches on the SI-$45^\circ$ and II-$0^\circ$ benchmarks, as demonstrated in the comparison with existing representative methods in Sec.~\ref{sec: comparison}. 

\begin{table}[htbp]
\centering
\footnotesize
\setlength{\tabcolsep}{4pt}
\caption{Cross-scene Performance on DVSpeaker (Training: SI-$0^\circ$)}
\label{tab:baseline_train10}
\begin{threeparttable}
\begin{tabular}{lccc}
\toprule
\multicolumn{1}{c}{\textbf{Testing conditions}} & \multicolumn{1}{c}{\textbf{SI-$45^\circ$}} & \multicolumn{1}{c}{\textbf{SI-$90^\circ$}} & \multicolumn{1}{c}{\textbf{II-$0^\circ$}} \\
\midrule
\multicolumn{1}{c}{Accuracy (\%)} & \multicolumn{1}{c}{71.68} & \multicolumn{1}{c}{3.64} & \multicolumn{1}{c}{75.94} \\
\bottomrule
\end{tabular}
\end{threeparttable}
\vspace{-1.4em}
\end{table}

\subsection{Comparison}
\label{sec: comparison}

As reviewed in Sec.~\ref{sec:introduction} and Sec.~\ref{sec:relatedworks}, there are several works about lip-based VSR, and most of them~\cite{chen2025dynamiclip,kuang2023lipauth,das2019lip} are RGB-based and evaluated mainly under matched-scene settings. Based on the reported results in Table~\ref{tab:lip_video_based_methods}, these systems can achieve near-perfect accuracy ($\approx$99\%, except~\cite{das2019lip}) on proprietary datasets of comparable scale. Due to the unavailability of source code and the lack of a public cross-scene benchmark in these studies, together with fundamental differences in sensing modality and acquisition protocols, a controlled quantitative comparison on DVSpeaker is infeasible. 
Nevertheless, NeuroLip already reaches saturated performance under matched scenes, achieving 100\% accuracy on DVSpeaker across both sufficient and insufficient illumination conditions (Table~\ref{tab:our_method_datasets}). Importantly, our main advantage lies in the strict cross-scene setting, where the performance of generic recognition architectures drops drastically. Therefore, we reproduce a broad set of publicly available video-based and event-based baselines under identical data splits and training recipes on DVSpeaker (Table~\ref{tab:lip_cross_scene_comparison}). Under cross-scene evaluation (Train: SI-$0^\circ$), NeuroLip outperforms the strongest reproduced competitors on SI-$45^\circ$ and II-$0^\circ$ by 19.82\% and 8.54\% accuracy, respectively, demonstrating that the proposed event-specific representation (TVE+SSE+PCR) substantially improves generalization to unseen viewpoints and illumination.

\begin{table}[!htbp]
\centering
\footnotesize  
\setlength{\tabcolsep}{1.4pt}  
\renewcommand{\arraystretch}{1.05}
\caption{Reported Performance of Existing Lip-based VSR Methods}
\label{tab:lip_video_based_methods}
\begin{threeparttable}
\begin{tabular}{lcccccc}
\toprule
\textbf{Method} & \textbf{\#Subjects}  & \textbf{Mode} & \textbf{Backbone} & \textbf{Br.-$0^\circ$ } & \textbf{Br}.-$45^\circ$ & \textbf{Dim}-$0^\circ$ \\
\midrule
DynamicLip\cite{chen2025dynamiclip} & 50  & Dyn.+Sta. & CNN            & 99.06          & $\approx$98.50          & $\approx$98.50 \\
LipAuth\cite{kuang2023lipauth}      & 50  & Dyn.+Sta. & DTW            & 99.24          & --                     & 98.95 \\
Das et al.\cite{das2019lip}         & 35  & Sta.      & KNN            & 92.80           & --                     & -- \\
\bottomrule
\end{tabular}

\begin{tablenotes}[flushleft]
\footnotesize  
\item Mode: the extracted lip pattern. ``Dyn.'' refers to dynamic lip motion while ``Sta.'' refers to static lip physiological features; ``Dyn.+Sta.'' means both are used.
\item Br.: bright illumination; Dim: insufficient illumination. All accuracy values are reported by the original studies under their respective experimental illumination-angle conditions; Specifically, DynamicLip (Br.: 500~lux; Dim: 80~lux), LipAuth (Br.: 120--160~lux; Dim: 10--50~lux), Das et al. (not specified).
\item ``--'': Not reported/not applicable.
\end{tablenotes}
\end{threeparttable}
\end{table}

\begin{table*}[!tbp]
\centering
\footnotesize  
\setlength{\tabcolsep}{4pt}  
\renewcommand{\arraystretch}{1.05}
\caption{Comparison With Reproduced Related Works on the DVSpeaker Benchmark under Matched-scene and Cross-scene Conditions}
\label{tab:lip_cross_scene_comparison}
\begin{threeparttable}
\begin{tabular}{lccccccccccc}
\toprule
\multirow{2}{*}{\textbf{Method}} & \multirow{2}{*}{\textbf{\#Subjects}} & \multirow{2}{*}{\textbf{Mode}} & \multirow{2}{*}{\textbf{Backbone}} &
\multirow{2}{*}{\textbf{Representation}} &
\multicolumn{4}{c}{\textbf{Matched-scene}} &
\multicolumn{3}{c}{\textbf{Cross-scene (Train = SI-0$^\circ$)}} \\
\cmidrule(lr){6-9} \cmidrule(lr){10-12}
 &  &  &  &  &  
SI-$0^\circ$ & SI-$45^\circ$ & SI-$90^\circ$ & II-$0^\circ$ &
SI-$45^\circ$ & SI-$90^\circ$ & II-$0^\circ$ \\
\midrule
EgoEvGesture~\cite{wang2025egoevgesture} & 50  & Dyn.+Sta. & ResNet          & Event Surfaces  & 80.70  & 83.90      & 83.60  & 71.60  & 3.92  & 3.02  & 2.04 \\
MTGA~\cite{zhang2025mtga}               & 50  & Dyn.+Sta. & Transformer     & Frame+Graph     & 93.85 & 97.58     & 96.98 & 94.56 & 27.60  & 3.20   & 30.00 \\
SpikGRU2+\cite{dampfhoffer2024neuromorphic} & 50 & Dyn.+Sta. & Spiking GRU & Raw            & 88.40  & 87.50      & 94.40  & 85.50  & 23.70  & \textbf{5.60} & 3.10 \\
ANN~\cite{dampfhoffer2024neuromorphic}  & 50 & Dyn.+Sta. & BiGRU           & Frame          & 93.30  & 92.40      & 96.60  & 81.80  & 37.50  & 4.60   & 12.62 \\
Get~\cite{peng2023get}                  & 50 & Dyn.+Sta. & Transformer     & Group Token    & \textbf{100.00} & \textbf{100.00} & 99.60 & 99.80 & 50.10 & 5.70 & 37.30 \\
MSTP~\cite{tan2022multi}                & 50 & Dyn.+Sta. & ResNet          & Frame          & 99.80  & 99.40      & 99.10  & 99.80  & 2.40   & 3.60   & 3.90 \\
EV-Gait~\cite{wang2021event}            & 50 & Dyn.+Sta. & CNN             & Frame          & \textbf{100.00} & \textbf{100.00} & \textbf{100.00} & \textbf{100.00} & 8.20 & 2.10 & 4.00 \\
EST~\cite{gehrig2019end}                & 50 & Dyn.+Sta. & VGG             & Spike Tensor   & \textbf{100.00} & 99.90      & \textbf{100.00} & \textbf{100.00} & 51.86 & 5.02 & 1.98 \\
\midrule
Uniformerv2~\cite{li2022uniformerv2}    & 50 & Dyn.+Sta. & Transformer     & Frame          & 35.00    & 51.90      & 40.60  & 25.60  & 9.50   & 4.50   & 4.30 \\
Tanet~\cite{liu2021tam}                 & 50 & Dyn.+Sta. & ResNet          & Frame          & 99.80  & 99.90      & 99.80  & 99.90  & 20.20  & 2.90   & 47.30 \\
Timesformer~\cite{bertasius2021space}   & 50 & Dyn.+Sta. & Transformer     & Frame          & 95.00    & 97.50      & 96.60  & 89.90  & 17.30  & 3.00     & 32.30 \\
Tin~\cite{shao2020temporal}             & 50 & Dyn.+Sta. & ResNet          & Frame          & \textbf{100.00} & \textbf{100.00} & 99.80 & 99.40 & 17.20 & 4.30 & 41.70 \\
Slowfast~\cite{feichtenhofer2019slowfast} & 50 & Dyn.+Sta. & ResNet        & Frame          & 99.90  & 99.80      & 99.20  & \textbf{100.00} & 50.20 & 3.90 & 28.40 \\
Slowonly~\cite{feichtenhofer2019slowfast} & 50 & Dyn.+Sta. & ResNet        & Frame          & \textbf{100.00} & \textbf{100.00} & 99.90 & \textbf{100.00} & 26.90 & 5.30 & 49.40 \\
Tsm~\cite{lin2019tsm}                   & 50 & Dyn.+Sta. & ResNet          & Frame          & 99.90  & 99.50      & 99.90  & 99.90  & 19.30  & 5.50   & 43.80 \\
R2plus1d~\cite{tran2018closer}          & 50 & Dyn.+Sta. & ResNet          & Frame          & \textbf{100.00} & 99.90 & \textbf{100.00} & \textbf{100.00} & 25.10 & 3.70 & 60.40 \\
Trn~\cite{zhou2018temporal}             & 50 & Dyn.+Sta. & Inception    & Frame          & 97.90  & 99.30      & 99.80  & 97.90  & 17.30  & 4.20   & 27.70 \\
C2d~\cite{wang2018non}                  & 50 & Dyn.+Sta. & ResNet          & Frame          & 99.80  & 98.90      & 99.10  & \textbf{100.00} & 20.30 & 5.10 & 67.40 \\
Tsn~\cite{wang2016temporal}             & 50 & Dyn.+Sta. & ResNet          & Frame          & \textbf{100.00} & \textbf{100.00} & \textbf{100.00} & \textbf{100.00} & 26.00 & 4.50 & 19.60 \\
C3d~\cite{tran2015learning}             & 50 & Dyn.+Sta. & CNN             & Frame          & \textbf{100.00} & \textbf{100.00} & \textbf{100.00} & \textbf{100.00} & 11.70 & 2.30 & 66.70 \\
\midrule
\textbf{NeuroLip (Ours)}                   & 50 & Dyn.+Sta. & ResNet          & Weighted Voxel          &
\textbf{100.00} & \textbf{100.00} & \textbf{100.00} & \textbf{100.00} &
\textbf{71.68} & 3.64 & \textbf{75.94} \\
\bottomrule
\end{tabular}

\end{threeparttable}
\end{table*}

To ensure a comprehensive and fair comparison, we reproduce 20 representative classification works with publicly available source code. These methods, spanning both video-based action recognition~\cite{li2022uniformerv2,liu2021tam,bertasius2021space,shao2020temporal,feichtenhofer2019slowfast,lin2019tsm,tran2018closer,zhou2018temporal,wang2018non,wang2016temporal,tran2015learning} and event-based tasks~\cite{wang2025egoevgesture,zhang2025mtga,dampfhoffer2024neuromorphic,peng2023get,tan2022multi,wang2021event,gehrig2019end}, are adapted to the DVSpeaker benchmark under both matched-scene and cross-scene settings.  Note that event-based classification methods are directly applicable in DVSpeaker, as they are designed to operate on event streams. To make DVSpeaker compatible with these video-based models, we convert the event streams into accumulated frames, fixing the temporal bin to 30 to approximate the frame rate of standard color videos.

The quantitative results, summarized in Table~\ref{tab:lip_cross_scene_comparison}, reveal distinct performance patterns between matched-scene and cross-scene protocols. Under matched-scene evaluation, a majority of methods achieve near-perfect performance across all four conditions (SI-$0^\circ$, SI-$45^\circ$, SI-$90^\circ$, II-$0^\circ$). This suggests that when training and testing distribution aligns, identity recognition is relatively straightforward given sufficient spatial or temporal cues. 

\begin{table*}[htbp]
\centering
\footnotesize
\setlength{\tabcolsep}{10pt}
\caption{Few-shot Domain Adaptation Performance (Pre-trained on SI-$0^\circ$)}
\label{tab:FewShot}
\begin{threeparttable}
\begin{tabular}{l*{10}{S[table-format=3.2]}}
\toprule
\textbf{Few-shot Setting} & \textbf{1} & \textbf{2} & \textbf{3} & \textbf{4} & \textbf{5} & \textbf{6} & \textbf{7} & \textbf{8} & \textbf{9} & \textbf{10} \\
\midrule
SI-$0^\circ$$\rightarrow$SI-$45^\circ$ & 83.99 & 92.92 & 97.10 & 96.58 & 94.97 & 98.64 & 98.84 & 96.59 & 96.31 & 97.91 \\
SI-$0^\circ$$\rightarrow$II-$0^\circ$   & 93.98 & 96.45 & 98.08 & 98.60 & 96.11 & 97.13 & 99.42 & 97.87 & 97.58 & 98.67 \\
\bottomrule
\end{tabular}
\end{threeparttable}
\vspace{-1.4em}
\end{table*}

However, cross-scene evaluation reveals substantial differences in robustness.
Most competing methods~\cite{wang2025egoevgesture,zhang2025mtga,dampfhoffer2024neuromorphic,tan2022multi,wang2021event,li2022uniformerv2,liu2021tam,bertasius2021space,shao2020temporal,lin2019tsm,zhou2018temporal,wang2016temporal} suffer severe performance degradation under unseen viewpoints or illumination, often exhibiting highly unbalanced performance across different domain shifts. In contrast, NeuroLip demonstrates superior and consistent generalization. When trained solely on SI-$0^\circ$, it maintains robust accuracy on the challenging SI-$45^\circ$ (71.68\%) and II-$0^\circ$ (75.94\%) benchmarks,
surpassing the best competing methods by significant margins of 19.82\% and 8.54\%, respectively. Given that NeuroLip employs a standard ResNet architecture comparable to many baselines, it is reasonable to infer that its superior cross-scene performance is not primarily driven by model complexity. Instead, we attribute these gains to our weighted voxel representation design, including TVE, SSE, and PCR, which collectively emphasize temporally stable behavioral cues and motion-direction information that are inherently less sensitive to appearance distortion. 

Despite these advancements, the extreme SI-$90^\circ$ condition remains a formidable challenge for all evaluated methods, including ours. The drastic geometric distortion and severe self-occlusion inherent in orthogonal viewpoints significantly disrupt observable lip motion patterns, leading to uniformly low performance across the board. This finding underscores that large-angle generalization remains an open and largely unsolved problem in lip-based VSR, necessitating further research into invariant representation learning across both event-based and RGB video-based approaches.

Additionally, we observe that video-based action recognition methods consistently achieve high accuracy under matched-scene conditions (with most exceeding 99\%). However, both video-based and event-based approaches suffer sharp performance degradation in cross-scene settings due to geometric distortion and feature distribution shift. Specifically, under the low-light condition (II-$0^\circ$), many generic event-based methods exhibit a significant performance drop. This is likely attributable to their sensitivity to increased sensor noise and the substantial reduction of valid events under low illumination, which complicates reliable identity inference in highly noisy regimes. In contrast, by leveraging a carefully designed polarity-sensitive spatiotemporal representation (TVE and SSE), our method effectively suppresses spurious noise while enhancing weak signals, thereby achieving superior robustness in these challenging scenarios.  

\subsection{Few-Shot Domain Adaptation}
In practice, collecting sufficient training data under all potential conditions is often costly or infeasible. While the cross-scene protocol evaluated in the previous sections represents a zero-shot domain generalization setting (where the target distribution is entirely unseen), practical deployment often allows for minimal calibration. To address the challenging zero-shot setting and enhance the cross-scene performance, we evaluate few-shot domain adaptation. It builds upon the cross-scene protocol and permits only a few target-scene samples for post-training adaptation, making it a practical extension of cross-scene evaluation. Specifically, we fine-tune the model pre-trained on SI-$0^\circ$ using a varying number of samples (K-shots) from two typical target conditions: SI-$45^\circ$ and II-$0^\circ$.

As shown in Table~\ref{tab:FewShot}, NeuroLip exhibits rapid adaptation capabilities across both target conditions. 
For SI-$45^\circ$, transitioning from zero-shot to one-shot yields an immediate absolute gain of 12.31\%, boosting accuracy to 83.99\%. With two shots, the accuracy exceeds 90\%. Performance continues to rise, peaking at 98.84\% with seven shots. When the number of shots increases to 8--10, the accuracy exhibits mild fluctuations without further consistent improvement.

For the SI-$0^\circ$$\rightarrow$II-$0^\circ$ setting, NeuroLip demonstrates even stronger few-shot adaptation performance. One-shot fine-tuning already achieves a high accuracy of 93.98\%, which is significantly higher than the one-shot performance on SI-$45^\circ$ (83.99\%), indicating NeuroLip's better intrinsic adaptation to insufficient illumination variations. Two-shot adaptation improves accuracy to 96.45\%, and three-shot adaptation further raises it to 98.08\%. The performance peaks at 99.42\% with seven shots, representing the highest accuracy across all K-shot settings for II-$0^\circ$. Similar to the SI-$45^\circ$ setting, 8--10 shots show slight fluctuations (97.58\%--98.67\%) without continuous gains.

These results demonstrate that NeuroLip offers a highly effective middle ground for deployment. By utilizing as few as 1--2 adaptation samples, the system can recover near-matched-scene performance on both SI-$0^\circ$$\rightarrow$SI-$45^\circ$ and SI-$0^\circ$$\rightarrow$II-$0^\circ$ conditions, providing a practical and cost-efficient solution that circumvents the need for exhaustive data collection from multiple conditions.

\subsection{Ablation Studies}
In this section, we present a comprehensive quantitative analysis to assess the impact of key factors on lip-motion-based VSR performance. Specifically, we investigate the contributions of individual framework components, the sensitivity to hyperparameters (PCR strength $\lambda$ and compressed channel dimension $C$), and the influence of spoken content and training sample size. 

\subsubsection{Impact of LTA, SSE, and PCR}
To examine the contribution of each key component in the proposed framework, we conduct a systematic evaluation on DVSpeaker.
The full model incorporates all proposed modules, including TVE with LTA, SSE, and PCR. We define three ablated variants to isolate the functional role of specific modules: 
\begin{itemize}
\item Model A (w/o LTA): Replaces the LTA in the TVE stage with a fixed, non-adaptive temporal kernel. This variant evaluates the importance of data-driven temporal weighting.
\item Model B (w/o SSE$\&$PCR): Removes both the SSE module and the PCR loss. This variant serves as a baseline to access the combined effect of spatial refinement and polarity preservation mechanism. 
\item Model C (w/o PCR): Retains SSE but removes PCR loss. Comparing Model C with the full model allows us to isolate the specific gain attributed to polarity-aware regularization.
\end{itemize}

\begin{table}[htbp]
\centering
\footnotesize
\setlength{\tabcolsep}{4pt}
\caption{Component-wise Analysis Results on DVSpeaker}
\label{tab:performance_comparison}
\begin{threeparttable}
\begin{tabular}{l*{3}{S[table-format=3.2]}}
\toprule
\multicolumn{1}{c}{\textbf{Model}} & \multicolumn{3}{c}{\textbf{Test Conditions (Accuracy \%)}} \\
\cmidrule(lr){2-4}
\multicolumn{1}{c}{\textbf{(trained on SI-$0^\circ$)}} & 
\multicolumn{1}{c}{\quad SI-$45^\circ$} &  
\multicolumn{1}{c}{\quad II-$0^\circ$} &  
\multicolumn{1}{c}{\quad SI-$0^\circ$} \\  
\midrule
Full model & 71.68 & 75.94 & 100.00 \\
Model A & 69.22 & 73.72 & 99.90 \\
Model B & 71.36 & 64.02 & 99.90 \\
Model C & 69.44 & 77.84 & 100.00 \\
\bottomrule
\end{tabular}
\end{threeparttable}
\end{table}

All models are trained exclusively on the SI-$0^\circ$ condition and evaluated on the matched SI-$0^\circ$ condition as well as the unseen SI-$45^\circ$ and II-$0^\circ$ conditions to test cross-scene generalization. As shown in Table~\ref{tab:performance_comparison}, the full model achieves the most balanced performance across evaluation conditions, indicating that the proposed components contribute in a complementary manner.
Notably, all models perform well under the matched SI-$0^\circ$ condition, as identity-relevant lip-motion patterns are consistently observed and the input distribution remains stable, rendering the task less sensitive to component differences.

Model~A exhibits degraded performance under cross-scene settings, 
suggesting that replacing LTA with a fixed kernel limits the model’s ability to adaptively capture identity-relevant temporal dynamics of lip motion.

Model~B achieves comparable performance under the SI-$45^\circ$ condition but suffers a significant degradation under II-$0^\circ$. 
In the absence of the SSE and PCR, fine-grained local lip details are insufficiently preserved. 
Under low illumination, the increased spurious noise further obscures these subtle structures, making them more susceptible to being overwhelmed. 
In contrast, sufficient illumination exhibits reduced noise, combined with TVE, still allows the model to exploit relatively stable behavioral cues, leading to improved performance under SI-$45^\circ$.

Model~C achieves the highest accuracy under II-$0^\circ$ 
yet underperforms under SI-$45^\circ$. 
This can be attributed to the removal of PCR.
Under low-illumination conditions, event cameras generate a substantially higher level of spurious events, whose polarities are largely dominated by sensor noise rather than meaningful motion cues. As a result, enforcing polarity reconstruction by PCR may introduce supervision signals that are weakly correlated with identity-related information.
Removing PCR effectively eliminates this noise-induced supervision, 
leading to improved performance under II-$0^\circ$. Conversely, under the SI-$45^\circ$ condition, the absence of polarity constraints becomes detrimental.
Polarity information encodes critical motion direction cues that remain relatively stable across viewpoint changes.
Without PCR, the model relies more heavily on the fine-grained spatial details emphasized by the SSE module. Since these spatial details are highly susceptible to geometric distortion under viewpoint change, the model’s discriminability degrades under SI-$45^\circ$, highlighting the essential role of PCR in stabilizing representations against viewpoint variations.

Overall, these results indicate that the full model achieves a favorable balance between discriminative power under matched conditions and robustness to cross-scene variations, benefiting from the complementary contributions of TVE, SSE, and PCR.

\subsubsection{Impact of PCR strength $\lambda$}
\begin{figure}[tb]
\centering
\includegraphics[width=1\linewidth]{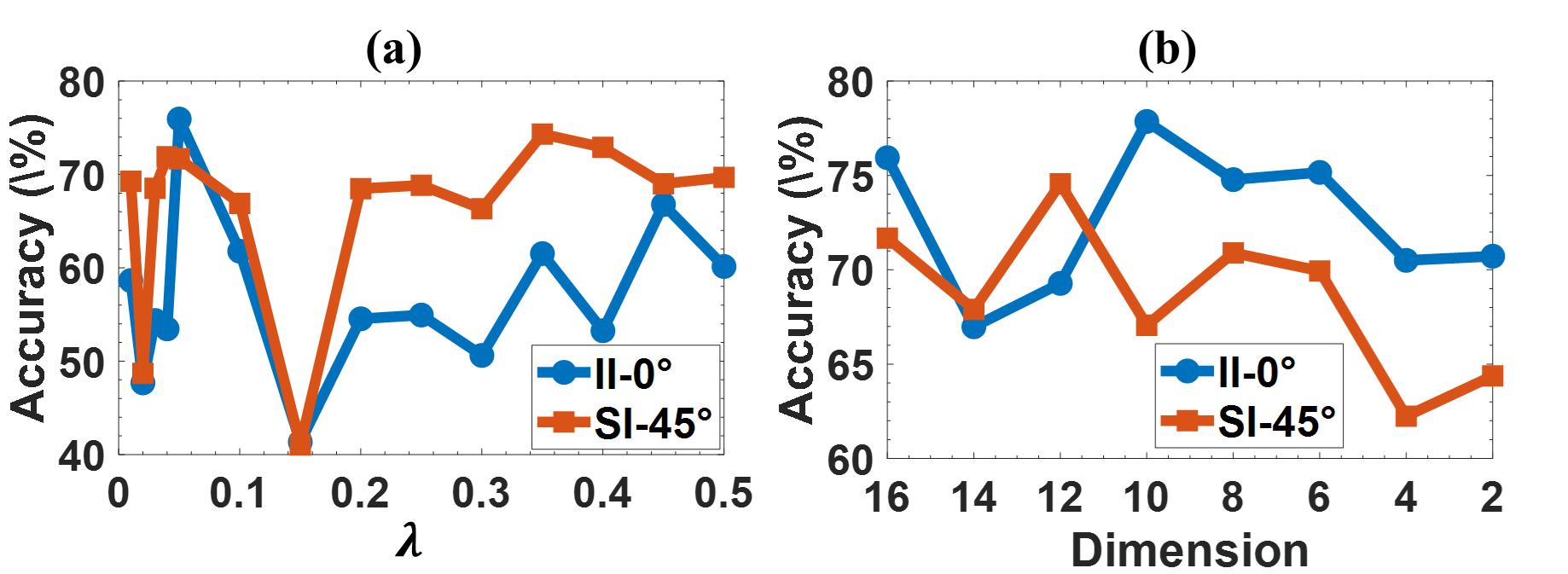} 
\caption{Impact of $\lambda$ (a) and $C$ in CCL (b).}
\label{combined_fig}
\vspace{-1.4em}
\end{figure}

The parameter $\lambda$ in Eq.~\eqref{TotalLoss} controls the strength of the $l_{\text{pcr}}$, with larger values enforcing stronger polarity constraints.
We investigate the impact of $\lambda$ in the range of 0.01 to 0.50. As shown in Fig.~\ref{combined_fig}(a), the recognition accuracy exhibits a clear non-monotonic trend.
As $\lambda$ increases from 0.01 to 0.05, performance improves steadily, peaking at 75.94\% on II-$0^\circ$ and 71.68\% on SI-$45^\circ$. However, pushing $\lambda$ beyond 0.05 results in a sharp drop in recognition accuracy between 0.05 and 0.15.
For larger values of $\lambda$, the behavior becomes unstable with partial recovery in certain conditions.
Notably, in this high-regularization regime, performance under II-$0^\circ$ consistently lags behind SI-$45^\circ$, corroborating our component-wise analysis that PCR is more critical for handling geometric viewpoint shifts rather than purely illumination-based degradation.

This phenomenon suggests a delicate balance in regularization strength. Insufficient regularization (small $\lambda$) fails to effectively reconstruct polarity signal,
whereas excessive regularization ($\lambda > 0.05$) forces the optimization to prioritize low-level polarity reconstruction over high-level identity discrimination, causing the model to overfit condition-specific patterns instead of learning identity-relevant and generalizable behavioral cues. Consequently, a moderate regularization strength ($\lambda = 0.05$) provides the best trade-off, enabling the network to leverage motion-direction cues for robust cross-scene generalization without compromising its primary discriminative objective.

\subsubsection{Impact of compressed channel dimension $C$ in CCL}
We further investigate the impact of compressed channel dimension ($C$) in the CCL, which controls the compactness of the learned event-based identity representation. We vary $C$ from 16 to 2, training on SI-0$^\circ$ and evaluating on SI-45 and II-0. The results shown in Fig.~\ref{combined_fig}(b) reveal that maintaining a sufficient channel capacity is crucial for cross-scene robustness. Specifically, the model achieves its peak performance at $C=16$ (75.94\% on II-$0^\circ$ and 71.68\% on SI-$45^\circ$). While some lower dimensions (e.g., $C=10$ for II-0$^\circ$, $C=12$ for SI-45$^\circ$) show isolated peaks, a general downward trend is observed as compression increases, particularly for $C$ less than 8. 
These results indicate that excessive channel compression discards critical spatiotemporal and polarity-related cues essential for identity discrimination. Therefore, a moderately compact channel configuration ($C=16$) is preferable to balance representation compactness with the preservation of fine-grained behavioral information.

\subsubsection{Impact of spoken contents}

To investigate the sensitivity of identity recognition to spoken content, we conduct a 5-fold cross-validation experiment under the matched SI-$0^\circ$ condition. As shown in Table~\ref{TextwiseExperiment_acc}, recognition accuracy varies across digits due to differences in articulatory characteristics. Digits such as `5' yield higher accuracy, likely because they involve larger and more sustained lip movements that produce clearer spatiotemporal event patterns. Conversely,  digits like `2' and `6' exhibit lower accuracy, as their articulation relies more on subtle, localized lip motion, resulting in weaker discriminative cues in the event stream.
These findings suggest a correlation between recognition robustness and the richness of articulation-induced motion: digits with more pronounced lip dynamics are easier to distinguish,
while those with subtle articulation pose greater difficulty. Nevertheless, NeuroLip maintains stable performance across all digits, 
demonstrating that while content influences difficulty, identity-related behavioral cues remain robust beyond specific phonetic patterns.

\begin{table}[tb]
\centering
\footnotesize
\setlength{\tabcolsep}{4pt}
\caption{Digit-wise Recognition Accuracy (\%) via 5-fold Cross-validation}
\label{TextwiseExperiment_acc}
\begin{threeparttable}

\begin{tabular}{c|S[table-format=3.2]S[table-format=3.2]S[table-format=3.2]S[table-format=3.2]S[table-format=3.2]|S[table-format=3.2]@{}c@{}S[table-format=3.2]}
\toprule
\textbf{Digit} & \textbf{Fold 1} & \textbf{Fold 2} & \textbf{Fold 3} & \textbf{Fold 4} & \textbf{Fold 5} & \multicolumn{3}{c}{\textbf{Mean $\pm$ Std}} \\
\midrule
0 & 95.00 & 98.00 & 95.00 & 98.00 & 97.00 & 96.60 & $\pm$ & 2.71 \\
1 & 91.00 & 97.00 & 100.00 & 95.00 & 100.00 & 96.60 & $\pm$ & 6.76 \\
2 & 84.00 & 85.00 & 89.00 & 92.00 & 94.00 & 88.80 & $\pm$ & 7.74 \\
3 & 85.00 & 94.00 & 90.00 & 94.00 & 92.00 & 91.00 & $\pm$ & 6.69 \\
4 & 97.00 & 86.00 & 92.00 & 96.00 & 93.00 & 92.80 & $\pm$ & 7.74 \\
5 & 97.00 & 99.00 & 99.00 & 100.00 & 99.00 & 98.80 & $\pm$ & 1.96 \\
6 & 86.00 & 89.00 & 94.00 & 97.00 & 77.00 & 88.60 & $\pm$ & 13.89 \\
7 & 93.00 & 96.00 & 97.00 & 98.00 & 92.00 & 95.20 & $\pm$ & 4.63 \\
8 & 88.00 & 96.00 & 93.00 & 95.00 & 97.00 & 93.80 & $\pm$ & 6.37 \\
9 & 91.00 & 92.00 & 98.00 & 99.00 & 96.00 & 95.20 & $\pm$ & 6.37 \\
\bottomrule
\end{tabular}
\begin{tablenotes}[flushleft]
\footnotesize
\item Mean $\pm$ Std: average accuracy and standard deviation over five folds.
\end{tablenotes}
\end{threeparttable}
\vspace{-1.4em}
\end{table}

\subsubsection{Impact of training sample size}
To investigate the impact of training sample sizes, we evaluate performance under the SI-$0^\circ$ matched-scene protocol by progressively increasing the number of training samples per subject. 
The sample size is increased in steps of 10, ensuring a balanced inclusion of all ten spoken digits (0--9) at each step.
As shown in Table~\ref{tab:SampleNum}, recognition accuracy increases rapidly at small sample sizes and then gradually saturates.
With only 10 samples (one per digit) per subject, the accuracy reaches 89.01\%.
Increasing this to 20 samples per subject boosts accuracy to over 98\%, and further increasing the sample size to 40 leads to near-perfect recognition.
This trend suggests that while a single sample per digit (10 samples) provides a strong baseline, it may be insufficient for the model to fully disentangle intrinsic identity cues from specific articulation contents. However, a moderate number of samples (e.g., 20 samples) proves sufficient for NeuroLip to robustly learn stable identity-related lip-motion patterns, demonstrating high data efficiency.

\begin{table}[htbp]
\centering
\footnotesize  
\setlength{\tabcolsep}{2pt}  
\caption{Impact of Training Sample Size }
\label{tab:SampleNum}  
\begin{threeparttable}
\begin{tabular}{l*{7}{S[table-format=3.2]}}
\toprule

\multicolumn{1}{c}{\textbf{Size}} & 
\multicolumn{1}{c}{10} & 
\multicolumn{1}{c}{20} & 
\multicolumn{1}{c}{30} & 
\multicolumn{1}{c}{40} & 
\multicolumn{1}{c}{50} & 
\multicolumn{1}{c}{60} & 
\multicolumn{1}{c}{70} \\
\midrule
Accuracy (\%) & 89.01 & 98.08 & 98.97 & 99.33 & 99.76 & 99.95 & 100.00 \\
\bottomrule
\end{tabular}
\end{threeparttable}
\end{table}

\section{Conclusion and Future Work}

This paper addresses the challenging problem of lip-motion-based VSR under a cross-scene protocol, where models trained in a single controlled environment must generalize to unseen viewpoints and illumination conditions.
To this end, we propose NeuroLip, an event-based lip-motion-driven identification framework that leverages the high temporal resolution and polarity-aware sensing of event cameras to capture robust, identity-specific behavioral patterns. The proposed NeuroLip integrates TVE, SSE, and PCR into a unified pipeline.
By emphasizing temporally stable behavioral cues rather than appearance-dependent features, the method achieves near-perfect performance under matched-scene conditions and significantly improved robustness under cross-scene evaluation.
Extensive experiments on our newly introduced DVSpeaker dataset demonstrate the effectiveness of the proposed NeuroLip under unseen illumination and viewpoint variations, while additional evaluation on the public DVSLip dataset verifies its generalization capability across different speaking content. Furthermore, comprehensive comparisons with reproduced existing methods on the DVSpeaker benchmark confirm that our proposed NeuroLip establishes a new performance standard for cross-scene VSR. 

As for future work, we plan to focus on three key directions:
First, extreme viewpoint variations remain challenging and may benefit from geometry-aware or view-invariant motion modeling.
Second, extending the framework to more unconstrained speaking content beyond digit articulation would improve practical applicability.
Finally, more adaptive polarity representation strategies may further enhance robustness under severe low-illumination noise.

\bibliography{references}

@STRING{IEEE_J_CASVT      = "{IEEE} Trans. Circuits Syst. Video Technol."}

@STRING{IEEE_J_IFS        = "{IEEE} Trans. Inf. Forensics Security"}

@STRING{IEEE_J_IOT        = "{IEEE} Internet Things J."}

@STRING{IEEE_J_MC         = "{IEEE} Trans. Mobile Comput."}

@STRING{IEEE_J_IP         = "{IEEE} Trans. Image Process."}

@STRING{IEEE_J_MM         = "{IEEE} Trans. Multimedia"}

@STRING{IEEE_J_AFFC       = "{IEEE} Trans. Affect. Comput."}

@STRING{IEEE_J_PAMI       = "{IEEE} Trans. Pattern Anal. Mach. Intell."}

@STRING{IEEE_J_SMCA       = "{IEEE} Trans. Syst., Man, Cybern. {A}"}

@STRING{IEEE_J_IM         = "{IEEE} Trans. Instrum. Meas."}

@article{yang2020preventing,
  title   = {Preventing Deepfake Attacks on Speaker Authentication by Dynamic Lip Movement Analysis},
  author  = {Yang, Chen-Zhao and Ma, Jun and Wang, Shilin and Liew, Alan Wee-Chung},
  journal = IEEE_J_IFS,
  volume  = {16},
  pages   = {1841--1854},
  year    = {2020}
}

@article{guo2022low,
  title   = {Low Cost and Latency Event Camera Background Activity Denoising},
  author  = {Guo, Shasha and Delbruck, Tobi},
  journal = IEEE_J_PAMI,
  volume  = {45},
  number  = {1},
  pages   = {785--795},
  year    = {2022}
}

@inproceedings{tan2022multi,
  title     = {Multi-grained Spatio-Temporal Features Perceived Network for Event-based Lip-Reading},
  author    = {Tan, Ganchao and Wang, Yang and Han, Han and Cao, Yang and Wu, Feng and Zha, Zheng-Jun},
  booktitle = {Proc. {IEEE/CVF} Conf. Comput. Vis. Pattern Recognit. ({CVPR})},
  pages     = {20062--20071},
  year      = {2022}
}

@article{Benedikt2010,
  title   = {Assessing the Uniqueness and Permanence of Facial Actions for Use in Biometric Applications},
  author  = {Benedikt, Lanthao and Cosker, Darren and Rosin, Paul L. and Marshall, David},
  journal = IEEE_J_SMCA,
  year    = {2010},
  volume  = {40},
  number  = {3},
  pages   = {449--460},
  month   = {May},
  doi     = {10.1109/TSMCA.2010.2041656}
}

@inproceedings{messer1999xm2vtsdb,
  title     = {{XM2VTSDB}: The Extended {M2VTS} Database},
  author    = {Messer, Kieron and Matas, Jiri and Kittler, Josef and Luettin, Juergen and Maitre, Gilbert and others},
  booktitle = {Proc. Int. Conf. Audio-Video-Based Biometric Person Authentication ({AVBPA})},
  pages     = {965--966},
  year      = {1999}
}

@inproceedings{movellan1994visual,
  title     = {Visual Speech Recognition with Stochastic Networks},
  author    = {Movellan, Javier},
  booktitle = {Proc. Adv. Neural Inf. Process. Syst. ({NeurIPS})},
  volume    = {7},
  year      = {1994}
}

@article{wright2020understanding,
  title   = {Understanding Visual Lip-Based Biometric Authentication for Mobile Devices},
  author  = {Wright, Carrie and Stewart, Darryl William},
  journal = {EURASIP J. Inf. Secur.},
  volume  = {2020},
  number  = {1},
  pages   = {3},
  year    = {2020}
}

@article{erzin2005multimodal,
  title   = {Multimodal Speaker Identification Using an Adaptive Classifier Cascade Based on Modality Reliability},
  author  = {Erzin, Engin and Yemez, Y{\"u}cel and Tekalp, A Murat},
  journal = IEEE_J_MM,
  volume  = {7},
  number  = {5},
  pages   = {840--852},
  year    = {2005}
}

@article{cao2014crema,
  title   = {{CREMA-D}: Crowd-sourced Emotional Multimodal Actors Dataset},
  author  = {Cao, Houwei and Cooper, David G and Keutmann, Michael K and Gur, Ruben C and Nenkova, Ani and Verma, Ragini},
  journal = IEEE_J_AFFC,
  volume  = {5},
  number  = {4},
  pages   = {377--390},
  year    = {2014}
}

@inproceedings{chung2016lip,
  title     = {Lip Reading in the Wild},
  author    = {Chung, Joon Son and Zisserman, Andrew},
  booktitle = {Proc. Asian Conf. Comput. Vis. ({ACCV})},
  pages     = {87--103},
  year      = {2016}
}

@article{cooke2006audio,
  title   = {An Audio-Visual Corpus for Speech Perception and Automatic Speech Recognition},
  author  = {Cooke, Martin and Barker, Jon and Cunningham, Stuart and Shao, Xu},
  journal = {J. Acoust. Soc. Am.},
  volume  = {120},
  number  = {5},
  pages   = {2421--2424},
  year    = {2006}
}

@article{koch2024one,
  title   = {One-shot Lip-Based Biometric Authentication: Extending Behavioral Features with Authentication Phrase Information},
  author  = {Koch, Brando and Grbi{\'c}, Ratko},
  journal = {Image Vis. Comput.},
  volume  = {142},
  pages   = {104900},
  year    = {2024}
}

@article{zue1990speech,
  title   = {Speech Database Development at {MIT}: {TIMIT} and Beyond},
  author  = {Zue, Victor and Seneff, Stephanie and Glass, James},
  journal = {Speech Commun.},
  volume  = {9},
  number  = {4},
  pages   = {351--356},
  year    = {1990}
}

@article{chen2025dynamiclip,
  title   = {{DynamicLip}: Shape-Independent Continuous Authentication via Lip Articulator Dynamics},
  author  = {Chen, Huashan and Xu, Yifan and Feng, Yue and Jian, Ming and Liu, Feng and Hu, Pengfei and Peng, Kebin and He, Sen and Wang, Zi},
  journal = {arXiv preprint arXiv:2501.01032},
  year    = {2025}
}

@article{kuang2023lipauth,
  title   = {{LipAuth}: Securing Smartphone User Authentication with Lip Motion Patterns},
  author  = {Kuang, Ling and Zeng, Fanzi and Liu, Daibo and Cao, Hangcheng and Jiang, Hongbo and Liu, Jiangchuan},
  journal = IEEE_J_IOT,
  volume  = {11},
  number  = {1},
  pages   = {1096--1109},
  year    = {2023}
}

@article{das2019lip,
  title   = {Lip Biometric Template Security Framework Using Spatial Steganography},
  author  = {Das, Srijan and Muhammad, Khan and Bakshi, Sambit and Mukherjee, Imon and Sa, Pankaj K and Sangaiah, Arun Kumar and Bruno, Andrea},
  journal = {Pattern Recognit. Lett.},
  volume  = {126},
  pages   = {102--110},
  year    = {2019}
}

@inproceedings{gehrig2019end,
  title     = {End-to-End Learning of Representations for Asynchronous Event-Based Data},
  author    = {Gehrig, Daniel and Loquercio, Antonio and Derpanis, Konstantinos G and Scaramuzza, Davide},
  booktitle = {Proc. {IEEE/CVF} Int. Conf. Comput. Vis. ({ICCV})},
  pages     = {5633--5643},
  year      = {2019}
}

@article{wang2025egoevgesture,
  title   = {{EgoEvGesture}: Gesture Recognition Based on Egocentric Event Camera},
  author  = {Wang, Luming and Shi, Hao and Yin, Xiaoting and Yang, Kailun and Wang, Kaiwei and Bai, Jian},
  journal = {arXiv preprint arXiv:2503.12419},
  year    = {2025}
}

@article{wang2021event,
  title   = {Event-Stream Representation for Human Gaits Identification Using Deep Neural Networks},
  author  = {Wang, Yanxiang and Zhang, Xian and Shen, Yiran and Du, Bowen and Zhao, Guangrong and Cui, Lizhen and Wen, Hongkai},
  journal = IEEE_J_PAMI,
  volume  = {44},
  number  = {7},
  pages   = {3436--3449},
  year    = {2021}
}

@inproceedings{zhang2025mtga,
  title     = {{MTGA}: Multi-view Temporal Granularity Aligned Aggregation for Event-Based Lip-Reading},
  author    = {Zhang, Wenhao and Wang, Jun and Luo, Yong and Yu, Lei and Yu, Wei and He, Zheng and Shen, Jialie},
  booktitle = {Proc. {AAAI} Conf. Artif. Intell. ({AAAI})},
  volume    = {39},
  number    = {10},
  pages     = {10176--10184},
  year      = {2025}
}

@inproceedings{dampfhoffer2024neuromorphic,
  title     = {Neuromorphic Lip-Reading with Signed Spiking Gated Recurrent Units},
  author    = {Dampfhoffer, Manon and Mesquida, Thomas},
  booktitle = {Proc. {IEEE/CVF} Conf. Comput. Vis. Pattern Recognit. ({CVPR})},
  pages     = {2141--2151},
  year      = {2024}
}

@inproceedings{peng2023get,
  title     = {{GET}: Group Event Transformer for Event-Based Vision},
  author    = {Peng, Yansong and Zhang, Yueyi and Xiong, Zhiwei and Sun, Xiaoyan and Wu, Feng},
  booktitle = {Proc. {IEEE/CVF} Int. Conf. Comput. Vis. ({ICCV})},
  pages     = {6038--6048},
  year      = {2023}
}

@inproceedings{tran2015learning,
  title     = {Learning Spatiotemporal Features with {3D} Convolutional Networks},
  author    = {Tran, Du and Bourdev, Lubomir and Fergus, Rob and Torresani, Lorenzo and Paluri, Manohar},
  booktitle = {Proc. {IEEE} Int. Conf. Comput. Vis. ({ICCV})},
  pages     = {4489--4497},
  year      = {2015}
}

@inproceedings{wang2018non,
  title     = {Non-local Neural Networks},
  author    = {Wang, Xiaolong and Girshick, Ross and Gupta, Abhinav and He, Kaiming},
  booktitle = {Proc. {IEEE} Conf. Comput. Vis. Pattern Recognit. ({CVPR})},
  pages     = {7794--7803},
  year      = {2018}
}

@inproceedings{tran2018closer,
  title     = {A Closer Look at Spatiotemporal Convolutions for Action Recognition},
  author    = {Tran, Du and Wang, Heng and Torresani, Lorenzo and Ray, Jamie and LeCun, Yann and Paluri, Manohar},
  booktitle = {Proc. {IEEE} Conf. Comput. Vis. Pattern Recognit. ({CVPR})},
  pages     = {6450--6459},
  year      = {2018}
}

@inproceedings{feichtenhofer2019slowfast,
  title     = {{SlowFast} Networks for Video Recognition},
  author    = {Feichtenhofer, Christoph and Fan, Haoqi and Malik, Jitendra and He, Kaiming},
  booktitle = {Proc. {IEEE/CVF} Int. Conf. Comput. Vis. ({ICCV})},
  pages     = {6202--6211},
  year      = {2019}
}

@inproceedings{liu2021tam,
  title     = {{TAM}: Temporal Adaptive Module for Video Recognition},
  author    = {Liu, Zhaoyang and Wang, Limin and Wu, Wayne and Qian, Chen and Lu, Tong},
  booktitle = {Proc. {IEEE/CVF} Int. Conf. Comput. Vis. ({ICCV})},
  pages     = {13708--13718},
  year      = {2021}
}

@inproceedings{bertasius2021space,
  title     = {Is Space-Time Attention All You Need for Video Understanding?},
  author    = {Bertasius, Gedas and Wang, Heng and Torresani, Lorenzo},
  booktitle = {Proc. Int. Conf. Mach. Learn. ({ICML})},
  year      = {2021}
}

@inproceedings{shao2020temporal,
  title     = {Temporal Interlacing Network},
  author    = {Shao, Hao and Qian, Shengju and Liu, Yu},
  booktitle = {Proc. {AAAI} Conf. Artif. Intell. ({AAAI})},
  volume    = {34},
  number    = {07},
  pages     = {11966--11973},
  year      = {2020}
}

@inproceedings{zhou2018temporal,
  title     = {Temporal Relational Reasoning in Videos},
  author    = {Zhou, Bolei and Andonian, Alex and Oliva, Aude and Torralba, Antonio},
  booktitle = {Proc. Eur. Conf. Comput. Vis. ({ECCV})},
  pages     = {803--818},
  year      = {2018}
}

@inproceedings{lin2019tsm,
  title     = {{TSM}: Temporal Shift Module for Efficient Video Understanding},
  author    = {Lin, Ji and Gan, Chuang and Han, Song},
  booktitle = {Proc. {IEEE/CVF} Int. Conf. Comput. Vis. ({ICCV})},
  pages     = {7083--7093},
  year      = {2019}
}

@inproceedings{wang2016temporal,
  title     = {Temporal Segment Networks: Towards Good Practices for Deep Action Recognition},
  author    = {Wang, Limin and Xiong, Yuanjun and Wang, Zhe and Qiao, Yu and Lin, Dahua and Tang, Xiaoou and Van Gool, Luc},
  booktitle = {Proc. Eur. Conf. Comput. Vis. ({ECCV})},
  pages     = {20--36},
  year      = {2016}
}

@article{li2022uniformerv2,
  title   = {{UniFormerV2}: Spatiotemporal Learning by Arming Image {ViTs} with Video {UniFormer}},
  author  = {Li, Kunchang and Wang, Yali and He, Yinan and Li, Yizhuo and Wang, Yi and Wang, Limin and Qiao, Yu},
  journal = {arXiv preprint arXiv:2211.09552},
  year    = {2022}
}

@article{tsuchihashi1974studies,
  title   = {Studies on Personal Identification by Means of Lip Prints},
  author  = {Tsuchihashi, Yasuo},
  journal = {Forensic Sci.},
  volume  = {3},
  pages   = {233--248},
  year    = {1974}
}

@article{chowdhury2022lip,
  title   = {Lip as Biometric and Beyond: A Survey},
  author  = {Chowdhury, Debbrota P and Kumari, Ritu and Bakshi, Sambit and Sahoo, Manmath N and Das, Abhijit},
  journal = {Multimed. Tools Appl.},
  volume  = {81},
  number  = {3},
  pages   = {3831--3865},
  year    = {2022}
}

@article{newby2025role,
  title   = {The Role of Facial Action Units in Investigating Facial Movements During Speech},
  author  = {Newby, Aliya A and Bhatta, Ambika and Kirkland III, Charles and Arnold, Nicole and Thompson, Lara A},
  journal = {Electronics},
  volume  = {14},
  number  = {10},
  pages   = {2066},
  year    = {2025}
}

@article{zhou2024securing,
  title   = {Securing Face Liveness Detection on Mobile Devices Using Unforgeable Lip Motion Patterns},
  author  = {Zhou, Man and Wang, Qian and Li, Qi and Zhou, Wenyu and Yang, Jingxiao and Shen, Chao},
  journal = IEEE_J_MC,
  volume  = {23},
  number  = {10},
  pages   = {9772--9788},
  year    = {2024}
}

@inproceedings{moreira2022neuromorphic,
  title     = {Neuromorphic Event-Based Face Identity Recognition},
  author    = {Moreira, Gon{\c{c}}alo and Gra{\c{c}}a, Andr{\'e} and Silva, Bruno and Martins, Pedro and Batista, Jorge},
  booktitle = {Proc. Int. Conf. Pattern Recognit. ({ICPR})},
  pages     = {922--929},
  year      = {2022}
}

@article{xiao2014facilitative,
  title   = {On the Facilitative Effects of Face Motion on Face Recognition and Its Development},
  author  = {Xiao, Naiqi G and Perrotta, Steve and Quinn, Paul C and Wang, Zhe and Sun, Yu-Hao P and Lee, Kang},
  journal = {Front. Psychol.},
  volume  = {5},
  pages   = {633},
  year    = {2014}
}

@article{wang2012physiological,
  title   = {Physiological and Behavioral Lip Biometrics: A Comprehensive Study of Their Discriminative Power},
  author  = {Wang, Shi-Lin and Liew, Alan Wee-Chung},
  journal = {Pattern Recognit.},
  volume  = {45},
  number  = {9},
  pages   = {3328--3335},
  year    = {2012}
}

@inproceedings{anina2015ouluvs2,
  title     = {{OuluVS2}: A Multi-view Audiovisual Database for Non-rigid Mouth Motion Analysis},
  author    = {Anina, Iryna and Zhou, Ziheng and Zhao, Guoying and Pietik{\"a}inen, Matti},
  booktitle = {Proc. {IEEE} Int. Conf. Autom. Face Gesture Recognit. ({FG})},
  volume    = {1},
  pages     = {1--5},
  year      = {2015}
}

@inproceedings{lu2018lippass,
  title     = {{LipPass}: Lip Reading-Based User Authentication on Smartphones Leveraging Acoustic Signals},
  author    = {Lu, Li and Yu, Jiadi and Chen, Yingying and Liu, Hongbo and Zhu, Yanmin and Liu, Yunfei and Li, Minglu},
  booktitle = {Proc. {IEEE} Conf. Comput. Commun. ({INFOCOM})},
  pages     = {1466--1474},
  year      = {2018}
}

@article{yang2025lip,
  title   = {{Lip-TWUID}: Noninvasive Through-Wall User Identification Using {SISO} Radar and Lip Movement Micro-Doppler Signatures with Limited Samples},
  author  = {Yang, Kai and Zhu, Dongsheng and Han, Chong and Guo, Jian and Sun, Suyun and Sun, Lijuan},
  journal = IEEE_J_IM,
  year    = {2025}
}

@article{lenz2022framework,
  title   = {A Framework for Event-Based Computer Vision on a Mobile Device},
  author  = {Lenz, Gregor and Picaud, Serge and Ieng, Sio-Hoi},
  journal = {arXiv preprint arXiv:2205.06836},
  year    = {2022}
}

@article{wang2025towards,
  title   = {Towards Mobile Sensing with Event Cameras on High-agility Resource-constrained Devices: A Survey},
  author  = {Wang, Haoyang and Guo, Ruishan and Ma, Pengtao and Ruan, Ciyu and Luo, Xinyu and Ding, Wenhua and Zhong, Tianyang and Xu, Jingao and Liu, Yunhao and Chen, Xinlei},
  journal = {arXiv preprint arXiv:2503.22943},
  year    = {2025}
}

@article{he2024lip,
  title   = {Lip Feature Disentanglement for Visual Speaker Authentication in Natural Scenes},
  author  = {He, Yi and Yang, Lei and Wang, Shilin and Liew, Alan Wee-Chung},
  journal = IEEE_J_CASVT,
  volume  = {34},
  number  = {10},
  pages   = {9898--9909},
  year    = {2024}
}

@article{cetingul2006discriminative,
  title   = {Discriminative Analysis of Lip Motion Features for Speaker Identification and Speech-Reading},
  author  = {Cetingul, H Ertan and Yemez, Y{\"u}cel and Erzin, Engin and Tekalp, A Murat},
  journal = IEEE_J_IP,
  volume  = {15},
  number  = {10},
  pages   = {2879--2891},
  year    = {2006}
}

@article{choras2010lip,
  title   = {The Lip as a Biometric},
  author  = {Chora{\'s}, Micha{\l}},
  journal = {Pattern Anal. Appl.},
  volume  = {13},
  number  = {1},
  pages   = {105--112},
  year    = {2010}
}

@inproceedings{lai2014sparse,
  title     = {Sparse Coding Based Lip Texture Representation for Visual Speaker Identification},
  author    = {Lai, Jun-Yao and Wang, Shi-Lin and Shi, Xing-Jian and Liew, Alan Wee-Chung},
  booktitle = {Proc. Int. Conf. Digit. Signal Process. ({DSP})},
  pages     = {607--610},
  year      = {2014}
}

@article{wrobel2017using,
  title   = {Using a Probabilistic Neural Network for Lip-Based Biometric Verification},
  author  = {Wrobel, Krzysztof and Doroz, Rafal and Porwik, Piotr and Naruniec, Jacek and Kowalski, Marek},
  journal = {Eng. Appl. Artif. Intell.},
  volume  = {64},
  pages   = {112--127},
  year    = {2017}
}

@inproceedings{niu2023lip,
  title     = {Lip Print Recognition Based on Convolutional Spiking Neural Network},
  author    = {Niu, Ben and Wang, Liu and Wu, Tianji and Zhang, Xiaofeng},
  booktitle = {Proc. Int. Conf. Image Signal Process. Pattern Recognit. ({ISPP})},
  volume    = {12707},
  pages     = {890--894},
  year      = {2023}
}

@inproceedings{luettin1996speaker,
  title     = {Speaker Identification by Lipreading},
  author    = {Luettin, Juergen and Thacker, Neil A and Beet, Steve W},
  booktitle = {Proc. Int. Conf. Spoken Lang. Process. ({ICSLP})},
  volume    = {1},
  pages     = {62--65},
  year      = {1996}
}

@article{zakeri2024whispernetv2,
  title   = {{WhisperNetV2}: {SlowFast} Siamese Network for Lip-Based Biometrics},
  author  = {Zakeri, Abdollah and Hassanpour, Hamid and Khosravi, Mohammad Hossein and Nourollah, Amir Masoud},
  journal = {arXiv preprint arXiv:2407.08717},
  year    = {2024}
}

@inproceedings{tran2017disentangled,
  title     = {Disentangled Representation Learning {GAN} for Pose-Invariant Face Recognition},
  author    = {Tran, Luan and Yin, Xi and Liu, Xiaoming},
  booktitle = {Proc. {IEEE} Conf. Comput. Vis. Pattern Recognit. ({CVPR})},
  pages     = {1415--1424},
  year      = {2017}
}

@inproceedings{huang2017beyond,
  title     = {Beyond Face Rotation: Global and Local Perception {GAN} for Photorealistic and Identity Preserving Frontal View Synthesis},
  author    = {Huang, Rui and Zhang, Shu and Li, Tianyu and He, Ran},
  booktitle = {Proc. {IEEE} Int. Conf. Comput. Vis. ({ICCV})},
  pages     = {2439--2448},
  year      = {2017}
}

@article{blanz2003face,
  title   = {Face Recognition Based on Fitting a {3D} Morphable Model},
  author  = {Blanz, Volker and Vetter, Thomas},
  journal = IEEE_J_PAMI,
  volume  = {25},
  number  = {9},
  pages   = {1063--1074},
  year    = {2003}
}

@inproceedings{hu2016face,
  title     = {Face Recognition Using a Unified {3D} Morphable Model},
  author    = {Hu, Guosheng and Yan, Fei and Chan, Chi-Ho and Deng, Weihong and Christmas, William and Kittler, Josef and Robertson, Neil M},
  booktitle = {Proc. Eur. Conf. Comput. Vis. ({ECCV})},
  pages     = {73--89},
  year      = {2016}
}

@inproceedings{tao2024gaitspike,
  title     = {{GaitSpike}: Event-Based Gait Recognition with Spiking Neural Network},
  author    = {Tao, Ying and Chang, Chip-Hong and Sa{\"\i}ghi, Sylvain and Gao, Shengyu},
  booktitle = {Proc. {IEEE} Int. Conf. {AI} Circuits Syst. ({AICAS})},
  pages     = {357--361},
  year      = {2024}
}

@article{petridis2017end,
  title   = {End-to-End Multi-View Lipreading},
  author  = {Petridis, Stavros and Wang, Yujiang and Li, Zuwei and Pantic, Maja},
  journal = {arXiv preprint arXiv:1709.00443},
  year    = {2017}
}

@inproceedings{lee2016multi,
  title     = {Multi-View Automatic Lip-Reading Using Neural Network},
  author    = {Lee, Daehyun and Lee, Jongmin and Kim, Kee-Eung},
  booktitle = {Proc. Asian Conf. Comput. Vis. ({ACCV})},
  pages     = {290--302},
  year      = {2016}
}

@article{ren2023spikepoint,
  title   = {{SpikePoint}: An Efficient Point-Based Spiking Neural Network for Event Cameras Action Recognition},
  author  = {Ren, Hongwei and Zhou, Yue and Huang, Yulong and Fu, Haotian and Lin, Xiaopeng and Song, Jie and Cheng, Bojun},
  journal = {arXiv preprint arXiv:2310.07189},
  year    = {2023}
}

@article{lagorce2016hots,
  title   = {{HOTS}: A Hierarchy of Event-Based Time-Surfaces for Pattern Recognition},
  author  = {Lagorce, Xavier and Orchard, Garrick and Galluppi, Francesco and Shi, Bertram E and Benosman, Ryad B},
  journal = IEEE_J_PAMI,
  volume  = {39},
  number  = {7},
  pages   = {1346--1359},
  year    = {2016}
}

@article{baldwin2022time,
  title   = {Time-Ordered Recent Event ({TORE}) Volumes for Event Cameras},
  author  = {Baldwin, R Wes and Liu, Ruixu and Almatrafi, Mohammed and Asari, Vijayan and Hirakawa, Keigo},
  journal = IEEE_J_PAMI,
  volume  = {45},
  number  = {2},
  pages   = {2519--2532},
  year    = {2022}
}

@article{chen2024egst,
  title   = {{EGST}: An Efficient Solution for Human Gaits Recognition Using Neuromorphic Vision Sensor},
  author  = {Chen, Liaogehao and Zhang, Zhenjun and Xiao, Yang and Wang, Yaonan},
  journal = IEEE_J_IFS,
  volume  = {19},
  pages   = {6144--6154},
  year    = {2024}
}

@article{gao2024hypergraph,
  title   = {Hypergraph-Based Multi-View Action Recognition Using Event Cameras},
  author  = {Gao, Yue and Lu, Jiaxuan and Li, Siqi and Li, Yipeng and Du, Shaoyi},
  journal = IEEE_J_PAMI,
  volume  = {46},
  number  = {10},
  pages   = {6610--6622},
  year    = {2024}
}

@inproceedings{wang2019space,
  title     = {Space-Time Event Clouds for Gesture Recognition: From {RGB} Cameras to Event Cameras},
  author    = {Wang, Qinyi and Zhang, Yexin and Yuan, Junsong and Lu, Yilong},
  booktitle = {Proc. {IEEE} Winter Conf. Appl. Comput. Vis. ({WACV})},
  pages     = {1826--1835},
  year      = {2019}
}

@article{spatiotemporal2011local,
  title   = {Local Ordinal Contrast Pattern Histograms for Spatiotemporal, Lip-Based Speaker Authentication},
  author  = {Chan, Chi-Ho and Goswami, Budhaditya and Kittler, Josef and Christmas, William J.},
  journal = IEEE_J_IFS,
  volume  = {7},
  number  = {2},
  pages   = {602--612},
  year    = {2012},
  doi     = {10.1109/TIFS.2011.2175920}
}

@inproceedings{tang2012deep,
  title={Deep Lambertian Networks},
  author={Tang, Yichuan and Salakhutdinov, Ruslan and Hinton, Geoffrey E.},
  booktitle={Proc. Int. Conf. Mach. Learn. (ICML)},
  pages={1419--1426},
  year={2012}
}

@article{tsai2021pam,
  title        = {{PAM}: Pose Attention Module for Pose-Invariant Face Recognition},
  author       = {Tsai, En-Jung and Yeh, Wei-Chang},
  journal      = {arXiv preprint arXiv:2111.11940},
  year         = {2021}
}

@article{wu2016comprehensive,
  title={A comprehensive study on cross-view gait based human identification with deep cnns},
  author={Wu, Zifeng and Huang, Yongzhen and Wang, Liang and Wang, Xiaogang and Tan, Tieniu},
  journal=IEEE_J_PAMI,
  volume={39},
  number={2},
  pages={209--226},
  year={2016},
}
\bibliographystyle{IEEEtran}

\newpage

\begin{IEEEbiography}[{\includegraphics[width=1in,height=1.25in,clip,keepaspectratio]{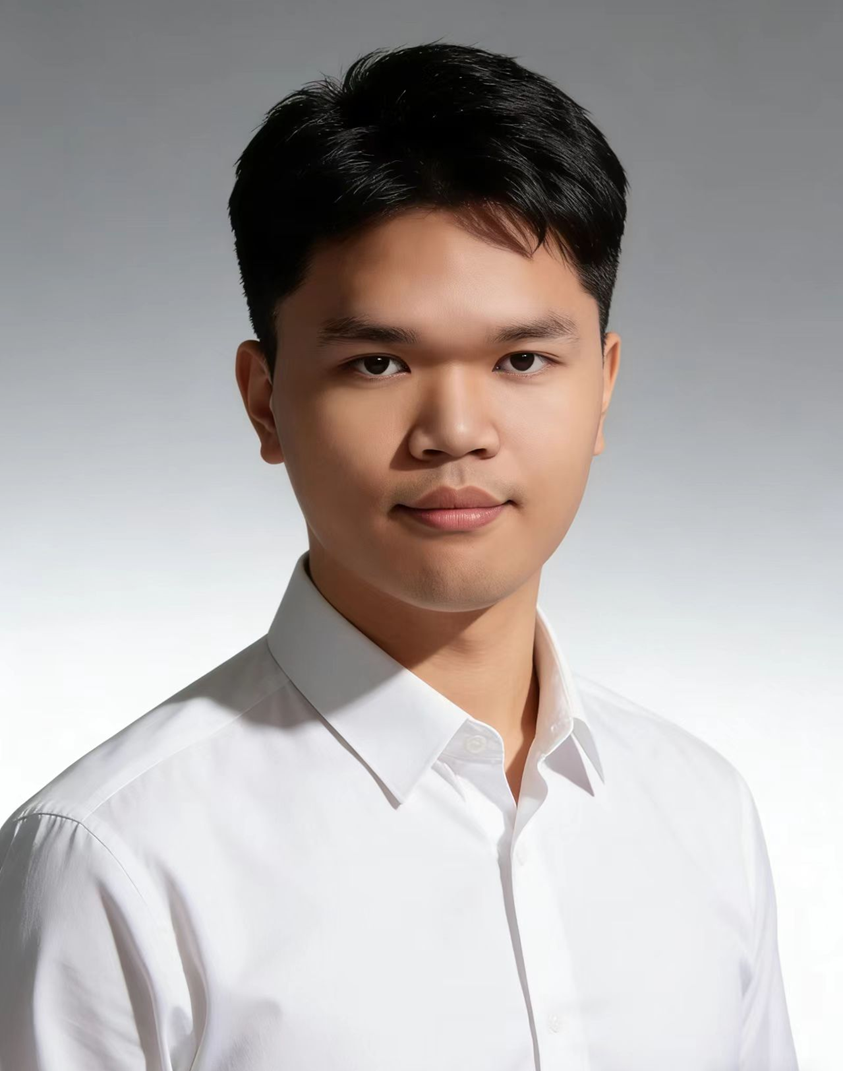}}]{Junguang Yao}
(S'26) received the B.E. degree from the College of Electronic Engineering, South China Agricultural University, Guangzhou, China, in 2021, and the M.S. degree from the College of Engineering, Southern University of Science and Technology, Shenzhen, China, in 2024. He is currently pursuing the Ph.D. degree with The Chinese University of Hong Kong, Shenzhen, China.

His research interests include behavioral biometrics and event cameras.
\end{IEEEbiography}

\begin{IEEEbiography}[{\includegraphics[width=1in,height=1.25in,clip,keepaspectratio]{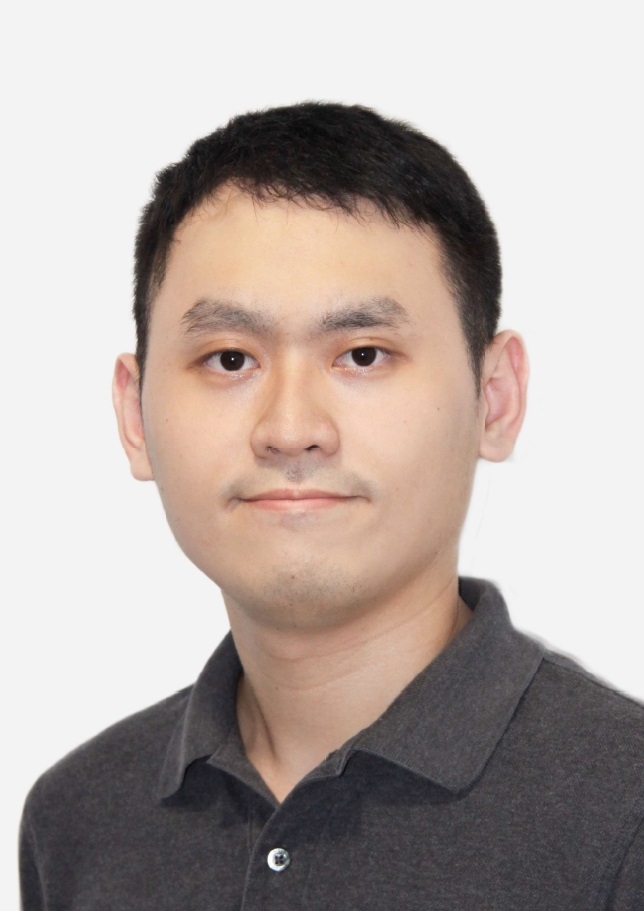}}]{Wenye Liu}
(S’17-M’23) received a B.S. degree in microelectronics from Shenzhen University, China, in 2014, a B.S. degree in physics from Umea University, Sweden, in 2014, an M.S. degree in IC design engineering from Hong Kong University of Science and Technology, and a Ph.D. degree from Nanyang Technological University (NTU), Singapore,
in 2021. He was a research fellow at NTU during 2021-2022. His current research interests include hardware security, machine learning accelerators and fault injection attack.
\end{IEEEbiography}

\begin{IEEEbiography}[{\includegraphics[width=1in,height=1.25in,clip,keepaspectratio]{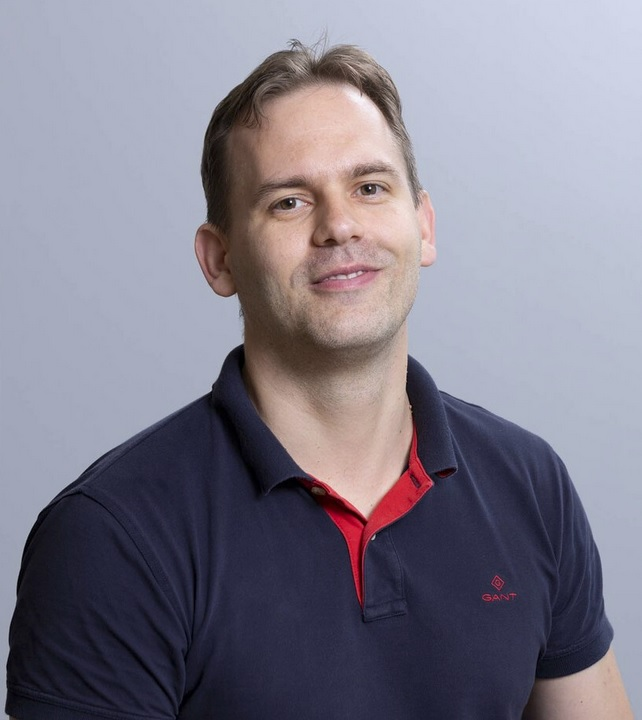}}]{Stjepan Picek} is a professor at the University of Zagreb, Croatia, and an associate professor at Radboud University, The Netherlands. Before that, Dr. Picek was an assistant professor at TU Delft, and a postdoctoral researcher at MIT, USA and KU Leuven, Belgium. His research interests are security/cryptography, machine learning, and evolutionary computation.\\
Up to now, Dr. Picek has given more than 70 invited talks and published more than 200 refereed papers. He is a program committee member and reviewer for a number of conferences and journals, as well as a member of several professional societies. Dr. Picek is a senior member of IEEE and an associate editor for several journals. He is a member of ELLIS and a Fellow of the Young Academy of Europe.
\end{IEEEbiography}

\begin{IEEEbiography}[{\includegraphics[width=1in, height=1.25in, clip, keepaspectratio]{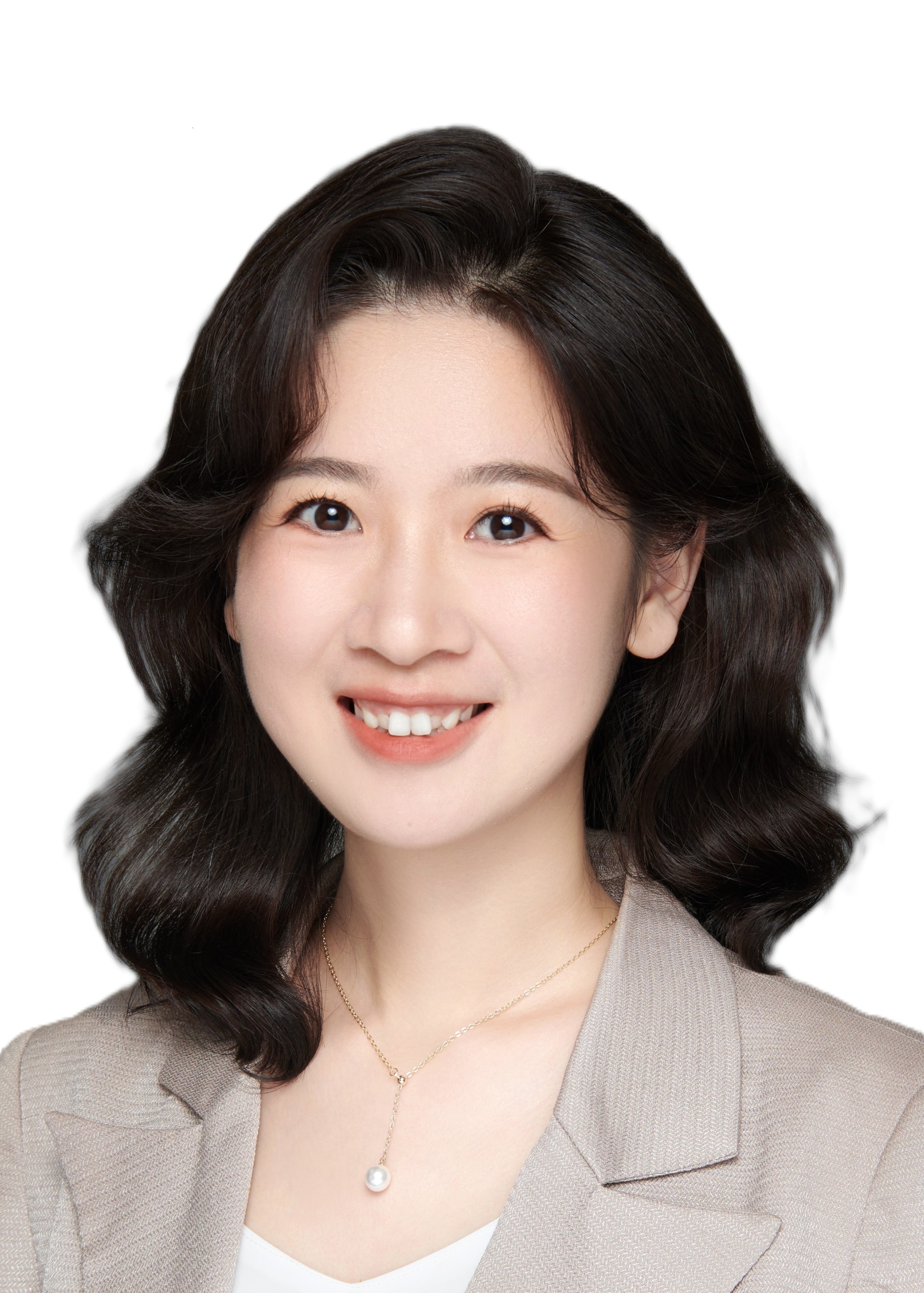}}]
{Yue Zheng}
(S'15-M'20) is currently an Assistant Professor at the Chinese University of Hong Kong, Shenzhen. Dr. Zheng received the B. Eng. degree from Shanghai University (SHU, China) and the Ph.D. degree from Nanyang Technological University (NTU, Singapore) in 2015 and 2020, respectively. Her areas of research include hardware security, security protocol design and trustworthy AI. Dr. Zheng is currently serving as an Associate Editor of IEEE Transactions on Information Forensics and Security, and EURASIP Journal on Information Security.

\end{IEEEbiography}

\end{document}